\documentclass{article} % For LaTeX2e
\usepackage{iclr2027_conference,times}

\usepackage{amsmath,amsfonts,bm}

\def\eqref#1{equation~\ref{#1}}
\def\1{\bm{1}}

\DeclareMathAlphabet{\mathsfit}{\encodingdefault}{\sfdefault}{m}{sl}
\SetMathAlphabet{\mathsfit}{bold}{\encodingdefault}{\sfdefault}{bx}{n}

\usepackage{algorithm}
\usepackage{algpseudocode}

\usepackage{pgfplots}
\usepackage{enumitem}
\usepackage{subcaption}

\usepackage[accsupp]{axessibility}  % Improves PDF readability for those with disabilities.
\usepackage[skins,breakable]{tcolorbox}
\tcbuselibrary{listings,breakable}
\tcbuselibrary{listings}%
\usepackage{lipsum}
\tcbset{listing engine={listings}}

\usepackage{xcolor}
\usepackage[dvipsnames]{xcolor}
\definecolor{myblue}{RGB}{66,134,218}
\definecolor{mygray}{RGB}{213,216,223}
\definecolor{mygreen}{RGB}{213,216,223}
\definecolor{incoming}{RGB}{206,206,206}
\definecolor{outgoing}{RGB}{186,233,95}
\definecolor{outgoingtop}{RGB}{227,242,221}
\definecolor{incomingtop}{RGB}{240,240,240}
\definecolor{pbutton}{RGB}{221,234,243} 
\definecolor{bottombg}{RGB}{217,217,217} 
\definecolor{bottomblue}{RGB}{86,117,142}

\usepackage{colortbl}
\definecolor{halfgray}{gray}{0.55}
\definecolor{ipython_frame}{RGB}{207, 207, 207}
\definecolor{deepblue}{rgb}{0,0,0.5}
\definecolor{deepred}{rgb}{0.6,0,0}
\definecolor{deepgreen}{rgb}{0,0.5,0}
\usepackage[skins,breakable]{tcolorbox}

\lstdefinelanguage[]{iPython}[]{python}{
    commentstyle=\color{cyan}\ttfamily,
    morekeywords=[3]{},
    morekeywords=[1]{self,__init__,_forward,GAB,Dict,Any,TensorType},
    keywordstyle=[3]\color{blue},
    stringstyle=\color{red}\ttfamily,
    keepspaces=true,
    showspaces=false,
    showstringspaces=false,
    frame=none,
    numbers=left,
    numberstyle=\scriptsize\color{halfgray},
    xleftmargin={0.3cm},
    basicstyle=\bf\fontfamily{cmtt}\scriptsize,
    keywordstyle=\color{deepgreen}\ttfamily,
    belowskip=-1pt,      % reduce space below
    aboveskip=-1pt,      % reduce space below
    xrightmargin=1pt,   % this sometimes works for frames*
    framexrightmargin=-1pt, % margin inside frame (right side), make it tiny
}

\usepackage{booktabs}
\usepackage{wrapfig}
\usepackage{listings}

\usepackage{hyperref}

\usepackage{orcidlink}
\usepackage{multirow}
\usepackage{arydshln}
\usepackage{colortbl}
\usepackage{url}

\title{LLM as a Tool, Not an Agent: Code-Mined Tree Transformations for Neural Architecture Search}

\author{\shortstack[l]{
Masakazu Yoshimura, Zitang Sun, Yuiko Sakuma,\\
Junji Otsuka, Atsushi~Irie \& Takeshi Ohashi
}\\
Sony Group Corporation\\
Tokyo, Japan \\
\texttt{\{masakazu.yoshimura, zitang.sun, yuiko.sakuma,}\\
\texttt{ junji.otsuka, atsushi.irie, takeshi.a.ohashi\}@sony.com}
}
\iclrfinalcopy % Uncomment for camera-ready version, but NOT for submission.
\begin{document}

\maketitle

\begin{abstract}
Neural Architecture Search (NAS) aims to automatically discover high-performing deep neural network (DNN) architectures. However, conventional algorithm-driven NAS relies on carefully hand-crafted search spaces to ensure executability, which restricts open-ended exploration. Recent coding-based agentic approaches using large language models (LLMs) reduce manual design, but current LLMs struggle to reliably generate complex, valid architectures, and their proposals are often biased toward a narrow set of patterns observed in their training data.
To bridge reliable algorithmic search with powerful LLM assistance, we propose LLMasTool, a hierarchical tree-based NAS framework for stable and open-ended model evolution. Our method automatically extracts reusable modules from arbitrary source code and represents full architectures as hierarchical trees, enabling evolution through reliable tree transformations rather than code generation.
At each evolution step, coarse-level planning is governed by a diversity-guided algorithm that leverages Bayesian modeling to improve exploration efficiency, while the LLM resolves the remaining degrees of freedom to ensure a meaningful evolutionary trajectory and an executable generated architecture. With this formulation, instead of fully agentic LLM approaches, our method explores diverse directions beyond the inherent biases in the LLM.
Our method improves over existing NAS methods by 0.69, 1.83, and 2.68 points on CIFAR-10, CIFAR-100, and ImageNet16-120, demonstrating its effectiveness.
\end{abstract}

\section{Introduction}

\begin{figure}[t]
\scalebox{0.97}{
\centering
\includegraphics[width=1.0\linewidth]{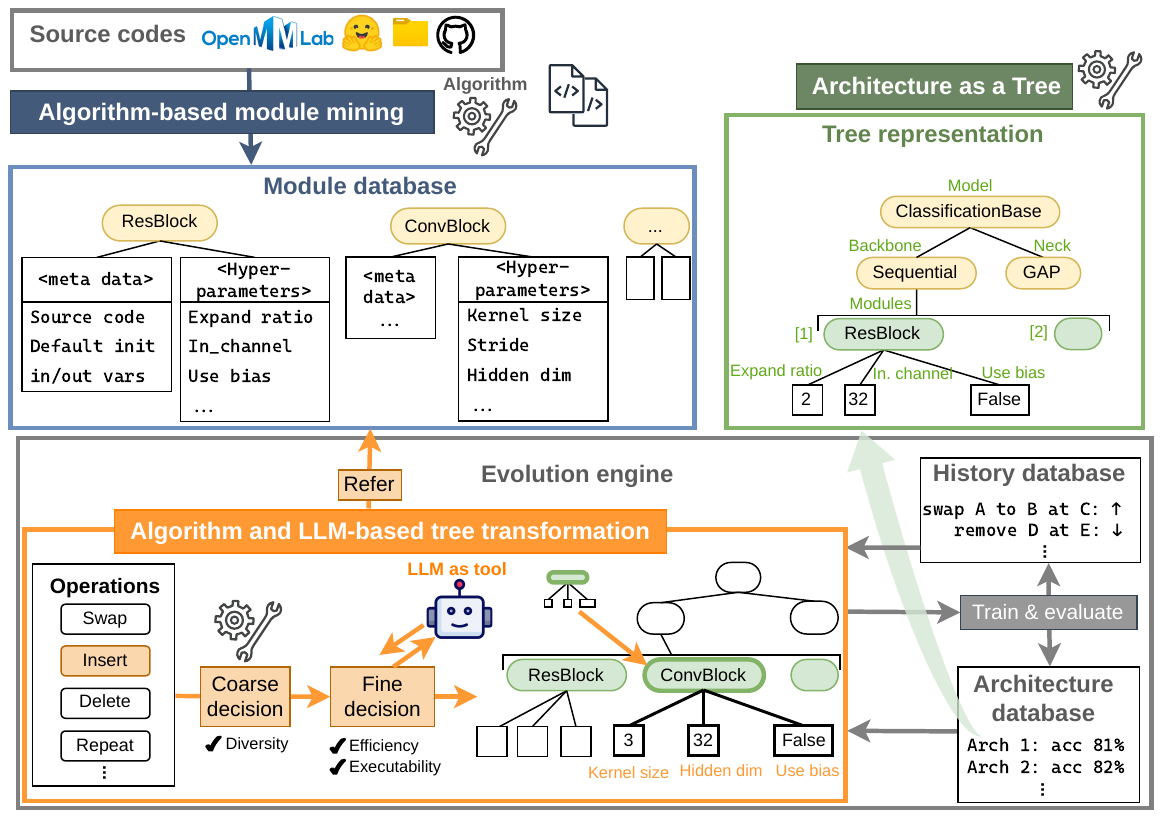}
}
\vspace{-10pt}
\caption{\textbf{Overview of the proposed NAS pipeline.} (Top-left) We mine reusable modules from source code to build a module database. (Top-right) architectures are represented as deploy-friendly hierarchical trees. (Bottom-left) an evolution engine applies controlled tree transformations, with an LLM serving as a tool for fine-grained refinement.}
\label{fig:overview}
\vspace{-10pt}
\end{figure}

Designing high-performing neural architectures remains a key driver of progress in deep learning, yet often requires substantial expertise and trial-and-error. Neural Architecture Search (NAS) aims to automate this process by exploring candidate architectures and optimizing their validation performance. Early NAS demonstrated that reinforcement learning~\citep{zoph2017nasrl} and evolutionary strategies~\citep{real2017evolution} can discover competitive architectures, while cell-based search enabled reusable designs across networks~\citep{zoph2018nasnet}. Subsequent work improved search efficiency through parameter sharing~\citep{pham2018enas}, differentiable optimization~\citep{liu2019darts}, and surrogate-guided search~\citep{liu2018pnas}. However, these methods typically rely on handcrafted search spaces and primitives to guarantee executability, restricting open-ended exploration and embedding strong designer priors~\citep{elsken2019nas_survey}.

Another line of work evolves existing architectures through structured edits. Network morphism reduces the cost of iterative modification by preserving network functions~\citep{wei2016network_morphism}, while Auto-Keras combines structured edits with Bayesian optimization~\citep{jin2019autokeras}. Hierarchical decompositions further factorize architecture search across abstraction levels~\citep{liu2019autodeeplab,qian2022nas_meets_trees}. Nevertheless, when edits become fine-grained across module choices, connectivity, and hyperparameters, the branching factor grows combinatorially, making useful feedback increasingly sparse~\citep{jin2019autokeras,russo2018thompson}.

Recently, large language models (LLMs) have emerged as a promising tool for NAS. LLMs can propose meaningful architectural modifications and abstract large low-level action spaces into semantically coherent decisions~\citep{chen2023evoprompting,zhou2025designprinciple}. However, LLM-driven NAS often suffers from limited controllability and difficulty guaranteeing executability, particularly when the LLM acts as an autonomous agent that generates architectures end-to-end~\citep{yang2025nader}. Furthermore, LLM proposals are inevitably influenced by patterns in their training data, which can limit exploration beyond familiar designs. These observations motivate a different paradigm: rather than asking an LLM to autonomously generate architectures, we use it as a \emph{tool} for localized decisions inside an algorithmically controlled and diversified search process.

In this work, we propose \textbf{LLMasTool}, an open-ended NAS framework that combines algorithmically diversity-guided evolution planning with LLM assistance. The key challenge is to provide a representation that is well-structured to be processed by algorithmic manipulation while expressive for LLM to understand. The source code representation is expressive for LLM but difficult to manipulate with algorithms. Existing NAS methods commonly use data-flow graphs, where features are represented as nodes and operations as edges. However, such representations are poorly aligned with how DNNs are implemented in source codes. In PyTorch~\citep{pytorch}, for example, models are expressed as hierarchically nested modules, with the data flow specified inside each module. Converting such implementations into data-flow graphs is non-trivial and requires intrusive rewriting.

We therefore represent architectures as \emph{deploy-friendly hierarchical trees}, whose nodes correspond to models, modules, components, and hyperparameters. We further develop an algorithmic module-mining procedure that extracts reusable modules from arbitrary PyTorch source code and stores them in a module database. This enables LLMasTool to directly reuse existing implementations and research knowledge rather than relying on LLM memorization.

On top of this representation, architecture evolution is formulated as a controlled tree transformation rather than free-form code generation. To prevent the search from being dominated by LLM priors, we employ a hierarchical coarse-to-fine decision process. A diversity-guided algorithm first determines the coarse direction of an edit, while the LLM resolves the remaining fine-grained choices. Historical outcomes are accumulated at the upper levels of this hierarchy, allowing Bayesian Thompson sampling to guide exploration and exploitation despite the sparsity of feedback in the open-ended search space.

Experiments on CIFAR-10, CIFAR-100, ImageNet16-120, and ImageNet-100 demonstrate the effectiveness of LLMasTool. The contributions are summarized as follows:
% \begin{itemize}
% \item We introduce \textbf{LLMasTool}, a hybrid NAS framework that represents complete architectures as deploy-friendly hierarchical trees, enabling precise and executable transformations while facilitating direct reuse of existing modules.
% \item We propose a \textbf{coarse-to-fine evolution strategy} that uses algorithmic mechanisms to diversify coarse-grained exploration and LLMs to guide fine-grained decisions, enabling controllable open-ended exploration beyond the LLM's priors.
% \item We introduce \textbf{hierarchical Thompson sampling} for efficient exploration in an open-ended search space.
% \end{itemize}
\begin{itemize}[itemsep=0pt, leftmargin=1.5em]
% \item A new perspective on LLM-assisted NAS. We argue that open-ended architecture search should not rely on an LLM as an autonomous architecture designer. Instead, we formulate NAS as an algorithm-driven exploration process in which the LLM acts only as a tool for resolving localized design decisions.
\item We formulate open-ended NAS as algorithm-driven exploration, with the LLM acting as a tool for localized design decisions rather than an autonomous architecture designer.
\item We introduce a deploy-friendly hierarchical tree representation together with algorithmic module mining from source code, allowing architecture evolution to be formulated as executable tree transformations rather than unrestricted code generation.
\item We propose a diversity-guided coarse-to-fine planning strategy and hierarchical Thompson sampling that effectively balance exploration and exploitation.
\item Extensive experiments across four datasets demonstrate the effectiveness of the proposed formulation, while ablation studies show that reducing reliance on the LLM can lead to better architecture discovery.
\end{itemize}
% \begin{itemize}[itemsep=0pt, leftmargin=1.5em]
% \item \textbf{A new perspective on LLM-assisted NAS.} We formulate open-ended NAS as algorithm-driven exploration, with the LLM acting as a tool for localized design decisions rather than an autonomous architecture designer.
% \item \textbf{A representation for controllable diversification.} We introduce a deploy-friendly hierarchical tree representation and algorithmic module mining from source code, formulating architecture evolution as executable tree transformations rather than unrestricted code generation.
% \item \textbf{A hierarchical exploration framework for sparse search spaces.} We propose diversity-guided coarse-to-fine planning with hierarchical Thompson sampling to balance exploration and exploitation while preserving executability.
% \item \textbf{Strong empirical validation.} Experiments on NAS-Bench-201 and ImageNet-100 show consistent improvements over existing LLM-based NAS methods, while ablations demonstrate that reducing LLM reliance can improve architecture discovery.
% \end{itemize}
\section{Related Work}
\vspace{-1mm}

\subsection{Neural Architecture Search (NAS)}
NAS optimizes neural network architectures in a data-driven manner. Most approaches represent architectures as data-flow graphs and optimize node types, hyperparameters, and connectivity. Reinforcement learning~\citep{zoph2017nasrl,zoph2018nasnet,pham2018enas} and evolutionary algorithms~\citep{suganuma2017genetic,liu2018pnas,so2019evolved} have been widely used for this purpose. Another major direction search within predefined super-networks through differentiable edge selection~\citep{liu2019darts,wu2019fbnet,cai2018proxylessnas,stamoulis2019single,zhang2024dependency} or partial weight and block sampling~\citep{yu2020bignas,cai2020once,zhao2021few,lu2023pa,jeon2025subnet,oh2025efficient,chen2021autoformer,chen2021searching,sakuma2024mixed}. GNN-based approaches have also been explored to improve architecture optimization efficiency~\citep{mills2023aio,salameh2023autogo,mills2024building}.

Although these methods can efficiently identify architectures within their search spaces, their predefined primitives and connectivity rules are mainly designed to ensure validity. This substantially restricts the diversity of architectures that can be explored.

\subsection{LLMs for NAS}
\vspace{-1mm}

Recent work has investigated LLMs for both conventional and open-ended NAS. GENEIUS~\citep{zheng2023can} uses an LLM to improve evolutionary search within a conventional search space, while LAPT~\citep{zhou2025design} uses an LLM to prune low-potential candidates. EvoPrompt~\citep{chen2023evoprompting} employs a coding LLM for mutation and crossover in open-ended NAS, and LLMatic~\citep{nasir2024llmatic} further combines LLM-based evolution with quality-diversity optimization. Genesys~\citep{cheng2025language} explores coding-based NAS for language models, while APD~\citep{kang2025revolutionizing} uses an LLM to propose zero-cost performance proxies. NADER~\citep{yang2025nader} extracts architectural ideas from recent papers and generates data-flow graphs rather than source code to reduce generation difficulty.

Despite their promise, LLM-based NAS methods still face challenges in controllability and executability, particularly when the LLM directly determines the complete architecture. Moreover, their exploration can remain strongly influenced by the LLM's prior knowledge. LLMasTool instead separates global algorithmic exploration from local LLM decision making, using an executable tree transformation framework to systematically explore a broader space.
\section{Methodology}
\vspace{-1mm}

We introduce a NAS framework that represents DNN architectures as hierarchical trees and performs architecture evolution through tree transformations, while using an LLM as a tool rather than an autonomous architecture generator. As illustrated in Fig.~\ref{fig:overview}, the framework consists of three components: (1) algorithmic module mining from existing source code, (2) a deploy-friendly hierarchical tree representation, and (3) diversity-guided coarse-to-fine evolution combining algorithmic decisions with LLM-assisted refinement.

\subsection{Module Mining from Existing Source Code}

We first construct a module database $\mathcal{D}_M$ from existing PyTorch source code. The mining procedure is entirely rule-based and does not rely on an LLM.

Given arbitrary PyTorch source codes, we parse it into an Abstract Syntax Tree (AST) and identify all classes inheriting from \texttt{torch.nn.Module}. Each extracted module is stored together with its search space and metadata:
$\mathcal{D}_M = {(m_i,s_i,\mu_i)}$,
where $m_i$, $s_i$, and $\mu_i$ denote a module, its tunable search space, and its metadata.

The constructor arguments of each module are treated as tunable search space, allowing the search to modify properties such as channel width, kernel size, and internal modules. We additionally record default argument values, the number of input/output variables of the \texttt{forward} method, and the relevant source code as its metadata. These metadata provide the information required to instantiate and transform modules during evolution.

Because module extraction is performed directly from source code using deterministic AST analysis, the resulting module database is reproducible and can incorporate implementations from arbitrary PyTorch-based repositories without manual intervention.

\subsection{Deploy-Friendly Tree Representation of DNN Architectures}
\label{sec:tree_definision}

\begin{figure}[t]
\centering
\includegraphics[width=1.0\linewidth]{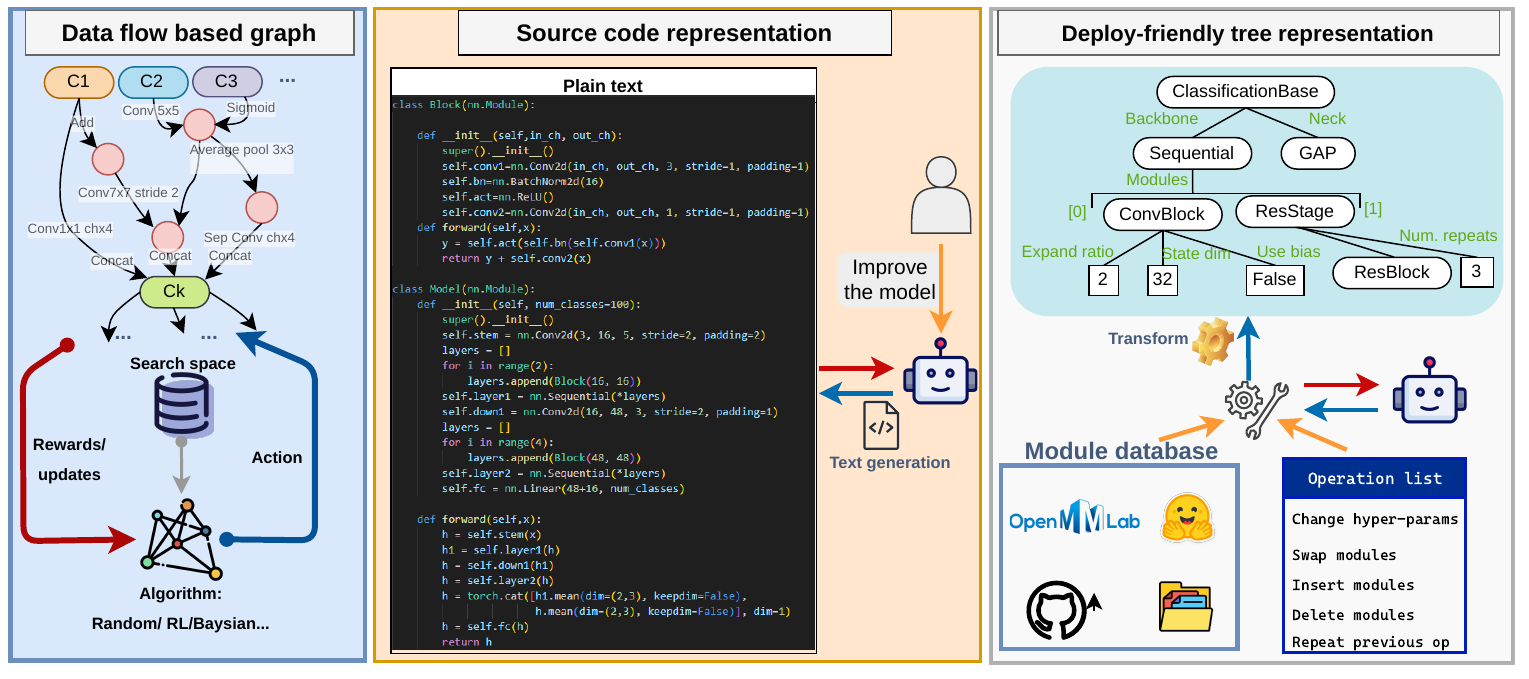}
\vspace{-20pt}
\caption{\textbf{Comparison of data-flow graph-, code-, and tree-based NAS representations.} Our deploy-friendly tree directly reflects the hierarchical structure of implemented DNN modules.}
\vspace{-10pt}
\label{fig:diff}
\end{figure}

Unlike conventional NAS methods that represent architectures as data-flow graphs~\citep{liu2019darts,liu2018pnas,mills2023aio,salameh2023autogo}, we represent an architecture as a deploy-friendly hierarchical tree, as shown in Fig.~\ref{fig:diff}. Higher-level nodes represent conceptual components such as the model and backbone, while lower-level nodes represent modules and their constructor arguments. Module nodes can contain further submodules, whereas hyperparameters form leaf nodes.

The tree is extended with list structures to support variable-length sequential compositions, such as stacking an arbitrary number of layers. This representation closely matches how DNNs are implemented in common frameworks and therefore enables direct reuse of existing source-code.

The hierarchical structure also naturally supports top-down refinement: a coarse architecture can first be selected and then progressively modified at lower levels, which is difficult with data-flow based graph representation.

\subsection{Architecture Evolution}

Our evolutionary search maintains a module database $\mathcal{D}_M$, an architecture database $\mathcal{D}_A$, and a history database $\mathcal{D}_H$.
The architecture database is defined as
$\mathcal{D}_A = \left\{ (A_i, e_i) \right\}_{i=1}^{N}$, where $A_i$ and $e_i$ denote the tree representation of a DNN architecture and the corresponding evaluation metric (e.g., accuracy).
At initialization, $\mathcal{D}_A$ contains at least one base architecture, which serves as the starting point of the evolution process.

At evolution step $t$, we sample a base architecture: $A_t^{base} \sim \mathcal{D}_A$,
and determine a tree transformation
\begin{equation}
 %T \sim p_{cf}\!\left( T \mid A_t^{base}, \mathcal{D}_A, \mathcal{D}_M, \mathcal{D}_H \right),
 T_t \sim p_{cf}\!\left( T \mid A_t^{base}, \mathcal{D}_A, \mathcal{D}_M, \mathcal{D}_H , \mathcal{LLM}\right),
 \label{eq:p_cf}
\end{equation}
where $\mathcal{D}_H = \left\{ (T_\tau, A_\tau^{base}, e_\tau-e_\tau^{base}) \right\}_{\tau=1}^{t-1}$ is a history database that stores previous experimental results with tree transformation $T_\tau$ and metric improvement $e_\tau-e_\tau^{base}$. The $p_{cf}$ represents the coarse-to-fine decision making discussed later.
The decided transformation is applied to $A_t^{base}$ as $A_t = T_t(A_t^{base})$.
The resulting architecture $A_t$ is evaluated by a feasibility function
\begin{equation}
\mathcal{E}(A_t^{base}, A_t, T_t) 
= \mathbf{1}_{\mathrm{exec}(A_t)}
\cdot 
\mathbf{1}_{\mathrm{const}(A_t)}
\cdot
\mathbf{1}_{\mathrm{intend}(A_t^{base}, A_t, T_t)},
\label{eq:epsilon}
\end{equation}
where $\mathbf{1}_{\mathrm{exec}(A_t)}$, $\mathbf{1}_{\mathrm{const}(A_t)}$, and $\mathbf{1}_{\mathrm{intend}(A_t^{base}, A_t, T_t)}$ verify whether $A_t$ is executable, $A_t$ satisfies target constraints such as FLOPs and parameter budgets, and the intended transformation $T_t$ is applied correctly.
If all feasibility checks are verified, we instantiate the architecture tree into an actual DNN, conduct training and evaluation, and then store the results in $\mathcal{D}_A$ and $\mathcal{D}_H$ as
\begin{equation}
\mathcal{D}_A \leftarrow \mathcal{D}_A \cup \{(A_t, e_t)\},\qquad \mathcal{D}_H \leftarrow \mathcal{D}_H \cup \{(T_t, A_t^{base}, e_t-e_t^{base})\}.
\label{eq:db_A}
\end{equation}
The overall evolution algorithm can be found in Algorithm \ref{alg:evolution}.

In the following, we describe the details about the proposed evolution with tree transformation $T$ and the coarse-to-fine decision making method $p_{cf}$.

\begin{figure}[t]
\centering
\includegraphics[width=1\linewidth]{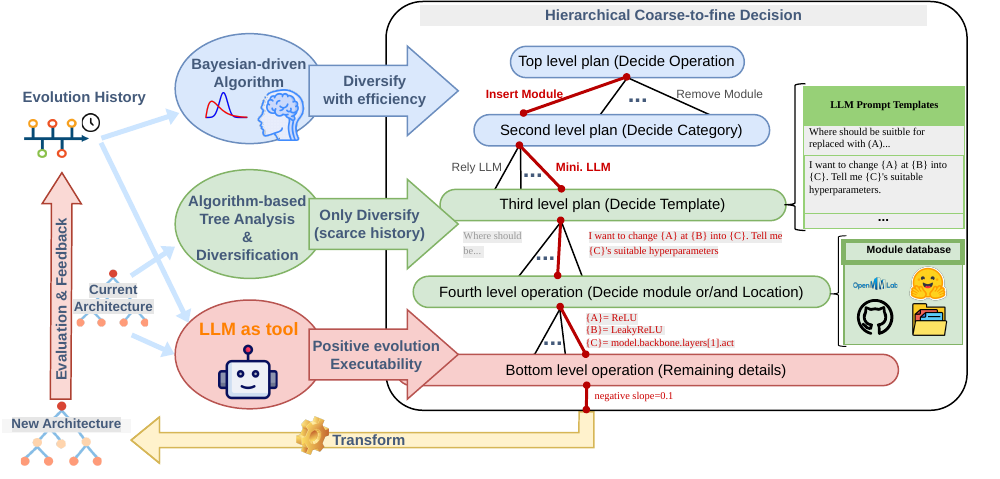}
\vspace{-20pt}
\caption{\textbf{Coarse-to-fine hierarchical decision process.} The high level balances exploration and exploitation using Bayesian sampling. The middle level enumerates candidate transformation directions algorithmically, while the low level uses the LLM to resolve the remaining degrees of freedom.}
\label{fig:c-to-f}
\vspace{-10pt}
\end{figure}

\subsubsection*{Tree Transformation as an Evolution Step}

We formulate architecture evolution as a tree transformation rather than source-code generation. This representation is compact, directly executable, and supports open-ended modifications. We define five operations for tree transformation:

\begin{tcolorbox}[
colback=gray!6,
colframe=gray!25,
boxrule=0.3pt,
arc=3pt,
left=8pt,right=8pt,top=3pt,bottom=3pt
]
\vspace{-5pt}
\begin{itemize}
\small
\item \textbf{Change Hyperparameter}: modify a leaf value, such as channel width or kernel size.
\item \textbf{Swap Module}: replace a sub-tree with another module from $\mathcal{D}_M$.
\item \textbf{Insert/Delete Module}: insert or remove a module at a list-structured position.
\item \textbf{Create Module}: construct a composite sub-tree from multiple existing sub-trees.
\item \textbf{Repeat Previous}: repeat a previously successful type of transformation.
\end{itemize}
\vspace{-10pt}
\end{tcolorbox}

The first four operations are the basic transformations $o\in\mathcal{O}_{basic}$, while \textbf{Repeat Previous} is a meta-operation that exploits previously successful directions. Given an operation, a concrete transformation requires determining where to edit and which module or value to use. The hierarchical tree makes these choices explicit and allows many of them to be handled algorithmically, while the LLM resolves the remaining semantic decisions. The detailed algorithms of each operation are provided in Sec. \ref{sec:app_operation_details}.

% The proposed tree-transformation-based evolution has four key advantages over the previous graph/coding-based NAS.
% (1) It directly reuses existing module assets, enabling reliable incorporation of prior knowledge without relying on LLM memorization.
% (2) Compared to a data-flow based graph representation, its hierarchical structure facilitates extraction and transplantation of high-performing sub-trees.
% (3) It supports rule-based controlled evolutions, allowing diverse and unbiased exploration.
% (4) A simple rule-based node-value check suffices to construct $\mathbf{1}_{\mathrm{intend}(A_t^{base}, A_t, T_t)}$, enabling a reliable history database and reducing unnecessary exploration.

\subsubsection*{Coarse-to-Fine Decision Making using LLM as a Tool}
At each evolution step, we must determine \textit{what} operation to apply, \textit{where} to apply it, and \textit{how} to instantiate it. Rather than delegating all decisions to the LLM, we factorize this process hierarchically and adopt a coarse-to-fine planning strategy combined with a rule-based algorithm as shown in Fig.~\ref{fig:c-to-f}.
The rule-based component determines the high-level structure of the transformation, while the LLM is used as a tool to determine a concrete tree transformation within the remaining degrees of freedom.
Specifically, our hierarchical sampling flow is as follows:
We \textbf{first} sample a tree transformation operation $o \sim p_1\!\left(o\right)$ from the basic tree transformation operations $o \in \mathcal{O}_{basic}$. For each operation, we prepare three categories of prompt templates $c \in \mathcal{C}$ (as shown in Fig. \ref{fig:c-to-f}): relying heavily on the prior knowledge of the LLM (\textit{rely LLM}), exploring opposed to the knowledge (inv. LLM), and focusing on diversity with minimal reliance on LLM (\textit{mini. LLM}).
The \textbf{second} sampling picks up one of these three categories $c \sim p_2\!\left(c \mid o \right)$, and the \textbf{third} sampling picks up a specific prompt template $\tau \in \mathcal{T}_c$ from the selected category $\tau \sim p_3\!\left(\tau \mid c, o \right)$.
As illustrated in Fig. \ref{fig:c-to-f}, each prompt template contains placeholders such as
\{A\} and \{B\}.
In the \textbf{fourth} sampling, to fill these placeholders, a rule-based algorithm retrieves all the corresponding node addresses or module names from the architecture tree $A_t^{base}$ and the module database $\mathcal{D}_M$. Then, it samples them as
$\phi 
\sim p_4\!\left(
\phi \mid c, o, \tau, A_t^{base}, \mathcal{D}_M)
\right)$ where $\quad \phi = (\mathrm{A},\mathrm{B}, ...)$.
The $p_1$ to $p_4$ have determined the coarse-grained aspects of the transformation at multiple levels of granularity. We then use LLM as a tool to create a complete transformation that generates a valid and executable architecture.
Specifically, we construct a complete prompt by combining the prompt template specified by $p_1$–$p_4$ with relevant metadata retrieved from the history and module databases. The resulting prompt is then provided to the LLM to generate the complete transformation $T$ according to $T \sim p_5(T\mid c,o,\tau,\phi, A_t^{base}, \mathcal{D}_M, \mathcal{D}_H)$. The details of the prompt templates and the resulting overall prompts are described in Sec. \ref{sec:prompt}.

In short, our coarse-to-fine decision making is
\begin{equation}
\footnotesize
p_{cf}(T\mid A_t^{base},\mathcal{D}_M,\mathcal{D}_H)=
\sum_{o,c,\tau,\phi}
p_5(T\mid c,o,\tau,\phi,\cdot)\,
p_4(\phi\mid c,o,\tau,\cdot)\,
p_3(\tau\mid c,o)\,
p_2(c\mid o)\,
p_1(o).
\end{equation} 
% where $\sum_{o,c,\tau,\phi}$ is the shorthand for
% $\sum_{o\in\mathcal{O}_{\text{basic}}}
%  \sum_{c\in\mathcal{C}}
%  \sum_{\tau\in\mathcal{T}_c}
%  \sum_{\phi\in\Phi(c,o,\tau;A_t^{base},\mathcal{D}_M)}$.
This decomposition has another merit. Although the complete transformation space is too large and sparse for each concrete decision to accumulate meaningful history, by grouping decisions hierarchically, useful feedback can accumulate at coarse levels, enabling efficient exploration while retaining fine-grained flexibility.

We implement coarse decisions ($p_1$ and $p_2$) using Bayesian Thompson sampling~\citep{russo2018thompson}. Since the magnitude of performance improvement can be noisy, we use only its sign:
$r_t=\mathbf{1}[e_t-e_t^{base}>0]$.
For each operation $o$, we maintain a success probability
$\theta_o\sim\mathrm{Beta}(\alpha_o,\beta_o)$, initialized with $\alpha_o=\beta_o=1$. After observing $r_t$, the corresponding posterior is updated as
\begin{equation}
\alpha_{o_t}\leftarrow\alpha_{o_t}+r_t,\qquad
\beta_{o_t}\leftarrow\beta_{o_t}+(1-r_t).
\end{equation}
Thompson sampling draws from each posterior to balance exploration and exploitation. Because the best operation may change as the architecture evolves, we further employ $\epsilon$-greedy Thompson sampling~\citep{do2024epsilon}, reserving probability $\epsilon$ for uniform exploration. The same procedure is applied to $p_2(c\mid o)$.

Thus, LLMasTool combines algorithmic diversity at coarse levels with LLM flexibility at fine levels, while hierarchical historical feedback makes open-ended exploration progressively more efficient.

\section{Experiments}
\begin{table*}[t]
\centering
\caption{Comparisons with state-of-the-art NAS methods on NAS-Bench-201 \cite{dong2020bench}-style setting. The $\diamondsuit$ indicates methods with parameter sharing, $\heartsuit$ indicates methods without parameter sharing, and $\bigstar$ are LLM-based methods. The $\pm$ in our results denotes the s.t.d. over three runs.
}
\label{tab:nasbench}
\scalebox{0.71}{\begin{tabular}{c|l|c|ll|ll|ll}
\toprule
 & \multirow{2}{*}{Method} & Search & \multicolumn{2}{c|}{CIFAR-10} & \multicolumn{2}{c|}{CIFAR-100} & \multicolumn{2}{c}{ImageNet16-120} \\
 & &  (archs) & validation & test & validation & test & validation & test \\
  \midrule
 & ~~~ResNet (our base arch.) &- & 90.83 & 93.97 & 70.42 & 70.86 & 44.53 & 43.63 \\
 & ~~~\textit{NAS-Bench-201 Optimal} &- & 91.61 & 94.37 & 73.49 & 73.51 & 46.77 & 47.31 \\ \midrule
\parbox{2.5mm}{\multirow{16}{*}{\rotatebox[origin=c]{90}{hand crafted search space}}} 
 & $\diamondsuit$ ENAS & - & 37.51$\pm$3.19 & 53.89$\pm$0.58 & 13.37$\pm$2.35 & 13.96$\pm$2.33 & 15.06$\pm$1.95 & 14.84$\pm$2.10 \\
 & $\diamondsuit$ DARTS  & - & 39.77$\pm$0.00 &  54.30$\pm$0.00 &  38.57$\pm$0.00 & 15.61$\pm$0.00 &  18.87$\pm$0.00 &  16.32$\pm$0.00 \\
 & $\diamondsuit$ SETN & - & 84.04$\pm$0.28 & 87.64$\pm$0.00 & 58.86$\pm$0.06 & 59.05$\pm$0.24 & 33.06$\pm$0.02 & 32.52$\pm$0.21  \\
 & $\diamondsuit$ DSNAS  & - & 89.66$\pm$0.29 & 93.08$\pm$0.13 &  30.87$\pm$16.40 & 31.01$\pm$16.38 & 40.61$\pm$0.09 & 41.07$\pm$0.09 \\
 & $\diamondsuit$ PC-DARTS & -  & 89.96$\pm$0.15 & 93.41$\pm$0.30 & 67.12$\pm$0.39 & 67.48$\pm$0.89 & 40.83$\pm$0.08 & 41.31$\pm$0.22 \\
 & $\diamondsuit$ SNAS & - & 90.10$\pm$1.04 & 92.77$\pm$0.83 & 69.69$\pm$2.39 & 69.34$\pm$1.98 & 42.84$\pm$1.79 & 43.16$\pm$2.64 \\ 
 & $\diamondsuit$ iDARTS & - & 89.86$\pm$0.60 & 93.58$\pm$0.32 &  70.57$\pm$0.24 & 70.83$\pm$0.48 &  40.38$\pm$0.59 & 40.89$\pm$0.68 \\
 & $\diamondsuit$ GDAS & - & 89.89$\pm$0.08 & 93.61$\pm$0.09 & 71.34$\pm$0.04 & 70.70$\pm$0.30 & 41.59$\pm$1.33 &  41.71$\pm$0.98  \\
 & $\diamondsuit$ DRNAS & - & {91.55$\pm$0.00} & 94.36$\pm$0.00  &  73.49$\pm$0.00 & 73.51$\pm$0.00 & 46.37$\pm$0.00 & 46.34$\pm$0.00  \\
 & $\diamondsuit$ $\beta$-DARTS & -  & {91.55$\pm$0.00} & 94.36$\pm$0.00  & 73.49$\pm$0.00 & 73.51$\pm$0.00 & 46.37$\pm$0.00 & 46.34$\pm$0.00  \\
 & $\diamondsuit$ $\Lambda$-DARTS & - & {91.55$\pm$0.00} & 94.36$\pm$0.00  &  73.49$\pm$0.00 & 73.51$\pm$0.00 & 46.37$\pm$0.00 & 46.34$\pm$0.00  \\ 
 & $\heartsuit$ REA & 500 &91.19$\pm$0.31 & 93.92$\pm$0.30& 71.81$\pm$1.12 & 71.84$\pm$0.99 & 45.15$\pm$0.89 & 45.54$\pm$1.03 \\
 & $\heartsuit$ RS & 500 & 90.93$\pm$0.36  & 93.70$\pm$0.36& 70.93$\pm$1.09 & 71.04$\pm$1.07& 44.45$\pm$1.10 & 44.57$\pm$1.25 \\
 & $\heartsuit$ REINFORCE & 500 & 91.09$\pm$0.37 & 93.85$\pm$0.37 & 71.61$\pm$1.12 & 71.71$\pm$1.09 & 45.05$\pm$1.02 & 45.24$\pm$1.18 \\
 & $\heartsuit$ BOHB & 500 & 90.82$\pm$0.53 & 93.61$\pm$0.52 & 70.74$\pm$1.29 & 70.85$\pm$1.28 & 44.26$\pm$1.36 & 44.42$\pm$1.49 \\
 & $\bigstar$ GENIUS & 10 & 91.07$\pm$0.20 & 93.79$\pm$0.09 & 70.96$\pm$0.33 & 70.91$\pm$0.72 & 45.29$\pm$0.81 & 44.96$\pm$1.02 \\ 
 & $\bigstar$ CoLLM-NAS & 100 & 91.59$\pm$0.04 & 94.37$\pm$0.01 & 73.44$\pm$0.12 & 73.44$\pm$0.15 & 46.62$\pm$0.10 & 46.79$\pm$0.28 \\
 % \rowcolor{blue!20}
 % & ~~~~\textit{Optimal} &- & 91.61 & 94.37 & 73.49 & 73.51 & 46.77 & 47.31 \\ \midrule
\midrule
 \parbox{2.5mm}{\multirow{6}{*}{\rotatebox[origin=c]{90}{open-ended NAS}}}
 & $\bigstar$ LLMatic & 2000 & \;\; - & 94.26$\pm$0.13 & \;\; - & 71.62$\pm$1.73 & \;\; - & 45.87$\pm$0.96 \\
 & $\bigstar$ LeMo-NADe(Gemini) & 30 & 82.94 & 81.76 & 52.12 & 52.96 & 30.34 & 31.02 \\
 & $\bigstar$ LeMo-NADe(GPT4) & 30 & 90.90 & 89.41 & 63.38 & 67.90 & 27.05 & 27.70 \\
 & $\bigstar$ NADER & 500 &  91.55 & 94.62 & 75.72 & 76.00 & 50.20 & \underline{50.52} \\
 \cdashline{2-9} % \hdashline
 & \multirow{2}{*}{$\bigstar$ LLMasTool} & 100 & 92.51$\pm$0.19& \underline{95.08$\pm$0.22}& \underline{76.48$\pm$0.35} & \underline{76.52$\pm$0.28} & \underline{50.40$\pm$0.22}& 50.43$\pm$0.55\\
 &  & 500 & \textbf{93.06}& \textbf{95.31}& \textbf{77.51} & \textbf{77.83} & \textbf{52.30}& \textbf{53.20}\\
 \bottomrule
 \end{tabular}}
\end{table*}

\subsection{Experimental Setup}
We evaluate LLMasTool under two experimental settings: (1) a NAS-Bench-201~\citep{dong2020bench}-style setting, following its training protocol and architectural scale on CIFAR-10, CIFAR-100~\citep{krizhevsky2009learning}, and ImageNet16-120~\citep{chrabaszcz2017downsampled} dataset; and (2) a larger-scale image size setting on ImageNet-100~\citep{tian2020contrastive}.

For NAS-Bench-201-style setting, we follow the protocol of prior work~\citep{yang2025nader}. Architectures are constrained to the maximum size of the benchmark search space, with a limit of 1.5M parameters and FLOP budgets of 0.2 GFLOPs for CIFAR-10/CIFAR-100 and 0.05 GFLOPs for ImageNet16-120. Although LLMasTool supports arbitrary initial architectures, we initialize the search from ResNet-32~\citep{he2016deep} following the prior work~\citep{yang2025nader}. Training follows the original NAS-Bench-201 protocol, including optimizer, epochs, and data splits. We report results after searching 100 and 500 architectures and use three runs when reporting standard errors.

For ImageNet-100, we use a budget of 2M parameters and 0.3 GFLOPs and primarily initialize from MobileNetV3-Small. Results with other initial architectures are provided in Appendix C.5.

For the module database, our algorithm extracts 167, 107, 467, and 681 modules from PyTorch v2.9.0~\citep{pytorch}, MMCV v2.1.0~\citep{githubOpenmmlabmmcv}, MMPretrain v1.2.0~\citep{githubOpenmmlabmmpretrain}, and timm v1.0.19~\citep{githubHuggingfacepytorchimagemodels}, respectively. We use Qwen3-8B~\citep{yang2025qwen3} as the LLM. Since our tree representation is similar to OpenMMLab configuration files~\citep{OpenMMLab}, we serialize the trees in an OpenMMLab-style format when prompting the LLM. Further details are provided in Appendix \ref{sec:app_exp_detail}.

\vspace{-3pt}
\subsection{Benchmarking}
Table~\ref{tab:nasbench} compares LLMasTool with representative NAS methods under the NAS-Bench-201-style setting. There are two categories of NAS methods: NAS methods operating on hand-crafted search spaces, including ENAS~\citep{pham2018enas}, DARTS~\citep{liu2019darts}, SETN~\citep{dong2019one}, PC-DARTS~\citep{Xu2020PC}, GDAS~\citep{dong2019searching}, DRNAS~\citep{chen2021drnas}, REA~\citep{real2019regularized}, random search (RS)~\citep{bergstra2012random}, REINFORCE~\citep{williams1992simple}, BOHB~\citep{falkner2018bohb}, GENIUS~\citep{zheng2023can}, and CoLLM-NAS~\citep{li2025collm}; and open-ended NAS methods, including LLMatic~\citep{nasir2024llmatic}, LeMo-NADe~\citep{rahman2024lemo}, and NADER~\citep{yang2025nader}.

Hand-crafted search spaces restrict exploration to a narrow set of manually designed architectures. Although recent methods such as $\Lambda$-DARTS~\citep{movahedi2022lambda} can approach the benchmark optimum efficiently, their discoveries remain bounded by the predefined topology.

Open-ended search removes this restriction but substantially increases the search difficulty. Code-generation methods such as LLMatic~\citep{nasir2024llmatic} and LeMo-NADe~\citep{rahman2024lemo} can generate diverse architectures but may suffer from instability and executability issues. NADER~\citep{yang2025nader} improves the reliability of LLM-based search by generating data-flow graphs and incorporating ideas from recent papers. However, its search remains dependent on the knowledge and representations available to the LLM.

LLMasTool substantially outperforms previous methods in both categories. From the hand-crafted setting, it improves test accuracy by 4.39 and 6.41 points on CIFAR-100 and ImageNet16-120, respectively. In open-ended NAS, it surpasses NADER by 1.83 and 2.68 points on the same datasets. These results demonstrate the effectiveness of the proposed formulation.

The improvement comes from how we formulate architecture evolution. Unlike methods that generate source code~\citep{nasir2024llmatic,rahman2024lemo} or data-flow graphs~\citep{yang2025nader}, LLMasTool generates neither the complete architecture nor an unrestricted graph. Instead, it performs explicit tree transformations. This allows coarse decisions to be diversified algorithmically, enables direct reuse of existing modules, and provides a structured representation for constructing complex architectures while maintaining executability.

\vspace{-1pt}
\subsection{Ablation Studies}
\begin{table*}[t]
\centering
\caption{Ablation Studies of the proposed components in the architecture evolution. We evolved from MobileNetV3-small0.5 to achieve better accuracy within one epoch training on ImageNet-100. The \textit{paper insp.} denotes the setting where 10 modification ideas are randomly selected and utilized from the 5,411 ideas that NADER \cite{yang2025nader} gathered from papers as inspiration. The grayed components are our proposal. 
}
\label{tab:ablation}
\scalebox{0.87}{
\begin{tabular}{cccccc|c}
 \toprule
 evo. step &\begin{tabular}{c}historical \\memory\end{tabular}&planning & \begin{tabular}{c}template \\category\end{tabular} &operation
&sampling&Acc.\\
 \midrule
 MobileNetV3  && -  & - &
-& - & 24.21 \\
 code-gen && LLM's idea & - &-
& uniform& 28.62\\
 config-gen && LLM's idea & - &
-& uniform&28.88\\
 \cellcolor{gray!30} tree-trans  && LLM's idea & - &\cellcolor{gray!30} basic& uniform&33.44\\
 \cellcolor{gray!30} tree-trans  &\checkmark& LLM's idea & - &
\cellcolor{gray!30} basic 
& uniform&33.45\\
 \cellcolor{gray!30} tree-trans  & \checkmark& paper inspi.& - & \cellcolor{gray!30} basic& uniform&34.98\\
 \cellcolor{gray!30} tree-trans  &\checkmark& LLM's idea & rely LLM &
\cellcolor{gray!30} basic 
& uniform&33.26\\
 \cellcolor{gray!30} tree-trans  &\checkmark& \cellcolor{gray!30} c-to-f & \cellcolor{gray!30} (rely+inv.) LLM&\cellcolor{gray!30} basic 
& uniform&34.00\\
 \cellcolor{gray!30} tree-trans  &\checkmark& \cellcolor{gray!30} c-to-f & \cellcolor{gray!30} (rely+inv.+w/o) LLM  &
\cellcolor{gray!30} basic& uniform&36.18\\
 \cellcolor{gray!30} tree-trans  &\checkmark& \cellcolor{gray!30} c-to-f & \cellcolor{gray!30} (rely+inv.+w/o) LLM &\cellcolor{gray!30} basic+repeat
& uniform&38.86\\
 \cellcolor{gray!30} tree-trans  & \checkmark& \cellcolor{gray!30} c-to-f & \cellcolor{gray!30} (rely+inv.+w/o) LLM & \cellcolor{gray!30} basic+repeat
& Thompson&33.22\\
 \cellcolor{gray!30} tree-trans  &\checkmark& \cellcolor{gray!30} c-to-f & \cellcolor{gray!30} (rely+inv.+w/o) LLM &
\cellcolor{gray!30} basic+repeat
& \cellcolor{gray!30} H-Thompson&\textbf{40.78}\\
 \bottomrule
\end{tabular}}
\vspace{-14pt}
\end{table*}
We perform ablation studies on ImageNet-100 to analyze the contribution of each component. Unless otherwise specified, we generate 100 architectures and train each for one epoch. The experimental setup and the longer-training experiments are provided in Appendices \ref{sec:app_imagenet_setup} and \ref{sec:app_imagenet100_100ep_exp}.

\paragraph{\textbf{Effectiveness of Module Mining.}}
\begin{wraptable}{r}{0.25\textwidth}
\vspace{-24pt}
\centering
\caption{Effectiveness of module mining}
\begin{tabular}{cc}
\toprule
modules & Acc. \\
\midrule
primitive & 33.21 \\
mined (ours) & \textbf{40.78} \\
\bottomrule
\end{tabular}
\label{tab:module}
\vspace{-15pt}
\end{wraptable}

We first compare our mined module database $\mathcal{D}_M$ with a primitive module database. Specifically, we compare (1) transformations using only MobileNetV3 modules and primitive modules implemented in the PyTorch library, and (2) transformations using the diverse modules automatically mined from existing source code.

As shown in Table~\ref{tab:module}, module mining substantially improves performance by enabling direct reuse of components developed in previous research. Since the procedure operates on arbitrary PyTorch source code, new implementations can also be incorporated without manual intervention.

\paragraph{\textbf{Effect of Tree Representation.}}
Next, we analyze the impact of representing architectures as a deploy-friendly hierarchical tree by comparing three architecture representations: \textbf{code-gen}, which directly generates or modifies source code; \textbf{config-gen}, which directly generates or modifies OpenMMLab-style configuration text; and \textbf{tree-trans (ours)}, which performs explicit transformations on the hierarchical tree. To isolate the effect of representation, all variants use the same instruction, ``please improve {architecture}'', together with the corresponding output format.

As can be seen by comparing the second to fourth rows of Table~\ref{tab:ablation}, code generation frequently produces familiar architectural patterns or introduces undesirable structural behavior, resulting in limited gains. Configuration generation still requires the LLM to determine the complete architecture, resulting in similar gains. In contrast, tree-trans explicitly specifies the transformation space and only asks the LLM to resolve the required local decisions, improving performance by 4.82 points over code-gen.
% Even without the full coarse-to-fine process ($p_2,p_3,p_4$), operation-level sampling through $p_1$ and the use of module database provide substantial gains. 
More importantly, the tree representation enables systematic control over modification locations and transformation types, which is essential for the coarse-to-fine planning proposal evaluated in the next section.

\paragraph{\textbf{Effect of Coarse-to-Fine Planning.}}

To evaluate the role of our coarse-to-fine decision making, we compare several planning strategies. As shown in Table~\ref{tab:ablation}, \textit{planning with LLM's idea}, where the LLM directly determines the entire transformation without hierarchical decomposition, tends to plateau beyond a certain performance level (The evolutionary process is visualized in Fig. \ref{fig:app_imnet100_scatter}). While methods such as NADER~\citep{yang2025nader} improve exploration diversity by extracting inspiration from multiple papers, such ideas remain bounded by the LLM's prior knowledge, limiting truly diverse exploration.

In contrast, our coarse-to-fine framework algorithmically enumerates all modifiable nodes, modification types, and available modules, then determines a general direction of change before using the LLM to resolve the remaining degrees of freedom. This enables diverse directional exploration while leveraging the LLM's prior knowledge to make effective and executable decisions within each direction.

For the template category, mini. LLM, which assigns minimal decision-making to the LLM, achieves the largest improvement. The result highlights the importance of our coarse-to-fine framework, which uses the LLM as a tool rather than an agentic planner.

\paragraph{\textbf{Effect of Meta-Level Operation.}}
We evaluate the \textbf{Repeat Previous} meta-operation.
As shown in Table \ref{tab:ablation}, adding this operation drastically improves the performance (The evolutionary lineage in Fig. \ref{fig:app_lineage} shows that this operation is effectively used to generate the best architecture.). 
Coarse-to-fine exploration guarantees diversity in the search process, but at the cost of substantially expanding the search space. Once a promising direction for improvement is identified, repeatedly exploring similar directions enhances the efficiency of the search.

\paragraph{\textbf{Effect of Hierarchical Thompson Sampling.}}
% We analyze the importance of hierarchical Thompson (H-Thompson) sampling, in which Thompson sampling is applied to both $p_1$ and $p_2$ under a hierarchical categorization scheme.
% We replace it with (1) uniform sampling, or (2) normal Thompson sampling, which treats each $p_3$ planning decision as an sampling item without structuring hierarchical categories.

% Uniform sampling lacks efficiency under the vast search space while normal Thompson sampling fails due to too sparse historical data.
% Our H-Thompson sampling provides a balance between exploration and exploitation, yielding both a faster improvement and better final architectures.
In Table \ref{tab:ablation}, we analyze the importance of hierarchical Thompson (H-Thompson) sampling, in which Thompson sampling is applied to both $p_1$ and $p_2$ under a hierarchical categorization scheme.
We replace it with (1) uniform sampling, or (2) normal Thompson sampling.
Uniform sampling lacks efficiency under the vast search space, while normal Thompson sampling fails due to too sparse historical data.
Our H-Thompson sampling provides a balance between exploration and exploitation, yielding both faster improvement (See also Fig. \ref{fig:app_imnet100_scatter}) and better final architectures.

\paragraph{\textbf{Summary of Findings.}}
Overall, the ablation studies demonstrate that the performance gain of LLMasTool does not arise from the LLM alone, but from the synergy between (1) rule-based module mining, (2) deploy-friendly tree representation, and (3) diversity-guided coarse-to-fine planning with Bayesian sampling.
These results validate the necessity of each design choice in enabling high-performance open-ended exploration.

\section{Conclusion}
We introduced LLMasTool, a NAS framework for open-ended architecture search. LLMasTool combines three key ideas: algorithmic module mining from existing source code, a deploy-friendly hierarchical tree representation, and coarse-to-fine evolution in which an LLM serves as a tool for fine-grained decisions rather than an autonomous architecture generator. The hierarchical decision process further enables Bayesian Thompson sampling to accumulate feedback at meaningful levels of abstraction. This combination balances the rule-based algorithm and LLM flexibility, achieving state-of-the-art performance among the compared open-ended NAS methods.

\subsection*{AI use statement}

In this work, we used generative AI tools for translation, grammar correction, and language polishing during manuscript preparation. We have not used generative AI tools to generate research ideas, develop the methodology, design experiments, analyze results, or draw scientific conclusions, and the remaining required disclosure tasks are not applicable to this work.
Additionally, we used generative AI tools to help implement a small portion of the source code. The majority of the source code was written by the authors, and all AI-assisted code was reviewed and understood by the authors, and the correctness was verified as part of the development and testing process.
We reviewed all AI-assisted work. The authors assume full responsibility for the final content of this work, including all text, claims, methodology, experimental results, and software artifacts produced with the aid of generative AI.

% \subsection*{Reproducibility statement}

% (This section is \textbf{recommended} and does not count toward the page limit.)

% It is important that the work published in ICLR is reproducible. Authors are
% strongly encouraged to include a paragraph-long Reproducibility Statement at the
% end of the main text (before references) to discuss the efforts that have been
% made to ensure reproducibility. This paragraph should not itself describe
% details needed for reproducing the results, but rather reference the parts of
% the main paper, appendix, and supplemental materials that will help with
% reproducibility. For example, for novel models or algorithms, a link to an
% anonymous downloadable source code can be submitted as supplementary materials;
% for theoretical results, clear explanations of any assumptions and a complete
% proof of the claims can be included in the appendix; for any datasets used in
% the experiments, a complete description of the data processing steps can be
% provided in the supplementary materials. Each of the above are examples of
% things that can be referenced in the reproducibility statement.

\bibliography{iclr2027_conference}
\bibliographystyle{iclr2027_conference}

\appendix
\section{Limitation and Future Works}
Various evaluations demonstrated that LLMasTool achieves significant performance improvements over previous work by enabling open-ended exploration without being constrained by handcrafted search spaces or the prior knowledge embedded in LLMs. However, several limitations remain.

In this work, we conducted proof-of-concept experiments using packages that include many implementations, such as PyTorch \citep{pytorch}, MMPretrain \citep{githubOpenmmlabmmpretrain}, MMCV \citep{githubOpenmmlabmmcv}, and timm \citep{githubHuggingfacepytorchimagemodels}. However, in recent years, other architectures, such as Mamba-based vision models \citep{zhu2024visionmamba, shi2024multiVmamba, xiao2025spatialmamba, yoshimura2026sf, yoshimuramambapeft}, have been proposed that are not included in these libraries. These models often achieve improvements by implementing custom CUDA kernels, and seamlessly integrating them into our framework would require a more sophisticated implementation that builds and manages multiple such CUDA kernels simultaneously.

Another limitation is the time required for the search process. The time required for LLM generation itself is about one minute, while training takes approximately 60 to 150 minutes for 200-epoch training on CIFAR-100 or ImageNet16-120. Therefore, compared to conventional NAS approaches based on evolutionary algorithms \citep{suganuma2017genetic, liu2018pnas, so2019evolved} or reinforcement learning \citep{zoph2017nasrl, zoph2018nasnet, pham2018enas}, the overall time overhead is minimal.
However, because the training cost for each candidate architecture is still high, searching over larger-scale model structures would require substantially more time.
To discover large-scale architectures, future work would benefit from zero-shot NAS methods that can reliably evaluate the relative quality of architectures without full training. The challenge of zero-shot NAS for open-ended exploration should be to accurately evaluate fundamentally different structures.

\section{Experimental Setup Details}
\label{sec:app_exp_detail}
\subsection{Common Setup}
Here, we describe the detailed experimental settings.
In our experiments, we conducted the architecture search with a budget of either 100 or 500 architecture trials.
To accelerate the experiments, the evolution process is not executed sequentially, but we run four evolution processes in parallel.

To balance the trade-off between exploration and exploitation, a promising base architecture is sampled from the architecture database $\mathcal{D}_A$ as $A_t^{base} \sim \mathcal{D}_A$.
Specifically, in the 100-trial setting, the base architecture is sampled from the top-5 architectures ranked by accuracy.
In the 500-trial setting, it is sampled from the top-25 architectures.
In terms of the coarse-to-fine decision making process, the $\epsilon$ for the $\epsilon$-greedy Thompson sampling \citep{do2024epsilon} is set to 0.5.

\subsection{NAS-Bench-201}
\begin{table}[h!]
\centering
\caption{The training hyperparemeter settings for CIFAR-10 and CIFAR-100}
\label{tab:setting_nasbench}
\begin{tabular}{l|l||l|l}
\hline
optimizer    & SGD        & initial LR    & 0.1                 \\
Nesterov     & \checkmark & ending LR     & 0                   \\
momentum     & 0.9        & LR schedule   & cosine              \\
weight decay & 0.0005     & epoch         & 200                 \\
batch size   & 256        & normalization & \checkmark          \\
random flip  & p=0.5      & random crop   & size=32, padding=4  \\
\hline
\end{tabular}
\end{table}
Following the experimental setup of NAS-Bench-201, we use the same partitions for the training, validation, and test datasets, as well as the same training hyperparameters. The hyperparameters for CIFAR-10 and CIFAR-100 are summarized in Table~\ref{tab:setting_nasbench}. For ImageNet16-120, we apply the same settings, except for a random crop size of 16 and a padding of 2 following the previous convention \citep{yang2025nader, liu2019darts}.

% \subsection{ImageNet-100 and ImageNet-1K}
\subsection{ImageNet-100}
\label{sec:app_imagenet_setup}
%\subsection{ImageNet-1K}
\begin{table}[h!]
\centering
\caption{The training hyperparemeter settings for ImageNet-100}
\label{tab:setting_imagenet1k}
\scalebox{0.9}{
\begin{tabular}{l|cc}
\hline
 & 1 epoch setup& 100 epoch setup\\
\hline
 batch size (total) & 64& 128\\
optimizer            & AdamW        & AdamW        
\\
optimizer momentum ($\beta_1$, $\beta_2$)& (0.9, 0.99)& (0.9, 0.99)
\\
learning rate
& 1e-3  
& 5e-3\\
minimal learning rate    
& 
1e-5
& 
5e-5\\
learning rate schedule & 
cosine      & cosine      
\\
weight decay         & 0.05     & 

0.05     
\\
 Warmup epochs & 0&5\\
random crop size   & 224              & 224              
\\
RandAug \cite{cubuk2020randaugment} & -& 
rand-m9-mstd0.5 
\\
 random erasing prob. & -& 0.25  
\\
mixup alpha     & -& 
0.8              
\\
 cutmix alpha        & -& 
1.0              \\
 label smoothing & -& 
0.1\\
 mixed precision (AMP)  & -& 
\checkmark  \\
\hline
\end{tabular}
}
\end{table}
 
% For ImageNet-1K, we follow the common Swin Transformer's hyperparameter settings \citep{liu2021swin} (Table~\ref{tab:setting_imagenet1k}) and use a 99\%/1\% train/validation split of the training data. For ImageNet-100, we use the predefined class subset in \citep{tian2020contrastive}. We apply the same training settings as ImageNet-1K, except for the batch size of 128 and the learning rate of 0.5$\times$.
For ImageNet-100, we use the predefined class subset and the train/validation split in \citep{tian2020contrastive}. We mainly follow the common Swin Transformer's hyperparameter settings for ImageNet-1K \citep{liu2021swin} but slightly adapted to the dataset size and the number of training epochs. In detail, we evaluated with one and 100 training epochs. For one epoch training setup, we use batch size of 64 and learning rate of 1e-3 without image augmentation. For 100 epoch training setup, we use batch size of 128, learning rate of 5e-4. More details can be found in Table~\ref{tab:setting_imagenet1k}.

\section{Additional Experiments}
\subsection{Analysis of the Evolutionary Process}

\begin{figure}
\centering
\scalebox{0.96}{
\hspace{10mm}
\begin{subfigure}{0.44\linewidth}
\hspace{-18mm}
% \begin{minipage}{0.48\linewidth}
\pgfplotsset{compat=1.18}
\centering
\begin{tikzpicture}
\begin{axis}[
    width=\linewidth,
    xlabel={Trials},
    ylabel={Accuracy},
    grid=major,
    ymin=20,
    ymax=42,
    xmin=0,
    xmax=100,
    scale only axis,
    legend style={font=\small, 
                  nodes={font=\scriptsize, scale=0.9, row sep=31pt},
                  },
    legend pos=north west
]

\addplot[
    only marks,
    red,
    mark=square*
] table[
    x=trial_id,
    y=acc,
    col sep=comma
]{1_fig/csv/in100_base.csv};
\addlegendentry{MobileNetV3-Small0.5}

\addplot[
    only marks,
    gray,
    mark=diamond*
] table[
    x=trial_id,
    y=acc,
    col sep=comma
]{1_fig/csv/in100_config_gen.csv};
\addlegendentry{w/o tree-trans. (config-gen)}

\addplot[
    only marks,
    purple,
    mark=*
] table[
    x=trial_id,
    y=acc,
    col sep=comma
]{1_fig/csv/in100_wo_thompson_repeat_minimum.csv};
\addlegendentry{w/o H-Thomp., repeat op., minimum LLM}

\addplot[
    only marks,
    green!60!black,
    mark=triangle*
] table[
    x=trial_id,
    y=acc,
    col sep=comma
]{1_fig/csv/in100_wo_thompson_repeat.csv};
\addlegendentry{w/o H-Thomp. and repeat op.}

\end{axis}
\end{tikzpicture}
\end{subfigure}
\hfill
\begin{subfigure}{0.44\linewidth}
% \end{minipage}
% \hfill
% \begin{minipage}{0.48\linewidth*0.5}
\pgfplotsset{compat=1.18}
\centering
\begin{tikzpicture}
\begin{axis}[
    width=\linewidth,
    xlabel={Trials},
    yticklabels=\empty,
    grid=major,
    ymin=20,
    ymax=42,
    xmin=0,
    xmax=100,
    scale only axis,
    legend style={font=\small, 
                  nodes={font=\scriptsize, scale=0.9, row sep=31pt},
                  },
    legend pos=north west
]

\addplot[
    only marks,
    red,
    mark=square*
] table[
    x=trial_id,
    y=acc,
    col sep=comma
]{1_fig/csv/in100_base.csv};
\addlegendentry{MobileNetV3-Small0.5}

\addplot[
    only marks,
    green!60!black,
    mark=triangle*
] table[
    x=trial_id,
    y=acc,
    col sep=comma
]{1_fig/csv/in100_wo_thompson_repeat.csv};
\addlegendentry{w/o H-Thomp. and repeat op.}

\addplot[
    only marks,
    blue,
    mark=diamond*
] table[
    x=trial_id,
    y=acc,
    col sep=comma
]{1_fig/csv/in100_wo_thompson.csv};
\addlegendentry{w/o H-Thompson}

\addplot[
    only marks,
    orange,
    mark=*
] table[
    x=trial_id,
    y=acc,
    col sep=comma
]{1_fig/csv/in100_proposed.csv};
\addlegendentry{proposed}

\end{axis}
\end{tikzpicture}
\end{subfigure}
% \end{minipage}
}
\caption{Evolutionary process on ImageNet-100 dataset with one epoch training.}
\label{fig:app_imnet100_scatter}

\end{figure}
Fig. \ref{fig:app_imnet100_scatter} shows the evolutionary process of the ablation study on ImageNet-100. First, when we simply let the LLM improve the OpenMMLab configuration file—which serves as the input and output format for our LLM—through a configuration generation approach (config-gen), the performance improvement quickly stagnates. In contrast, by adopting tree transformation as an evolutionary process while still allowing the LLM to perform substantial planning, the performance improves significantly.

Furthermore, by introducing prompt categories that minimize the use of the LLM in the planning stage, the initial speed of performance improvement becomes slower; however, the search is no longer constrained by the prior bias of the LLM, resulting in better performance gains in the later stages. In addition, introducing the repeat operation, which performs a similar transformation again when a performance improvement is observed, substantially accelerates the speed of performance improvement.

Finally, by employing hierarchical Thompson sampling, performance improvements accelerate in the later stages once a meaningful history has been accumulated. Although the process is highly stochastic and noisy, these results demonstrate that each proposed component contributes as intended.

\begin{figure}[ht]
    \centering
    \includegraphics[width=\linewidth]{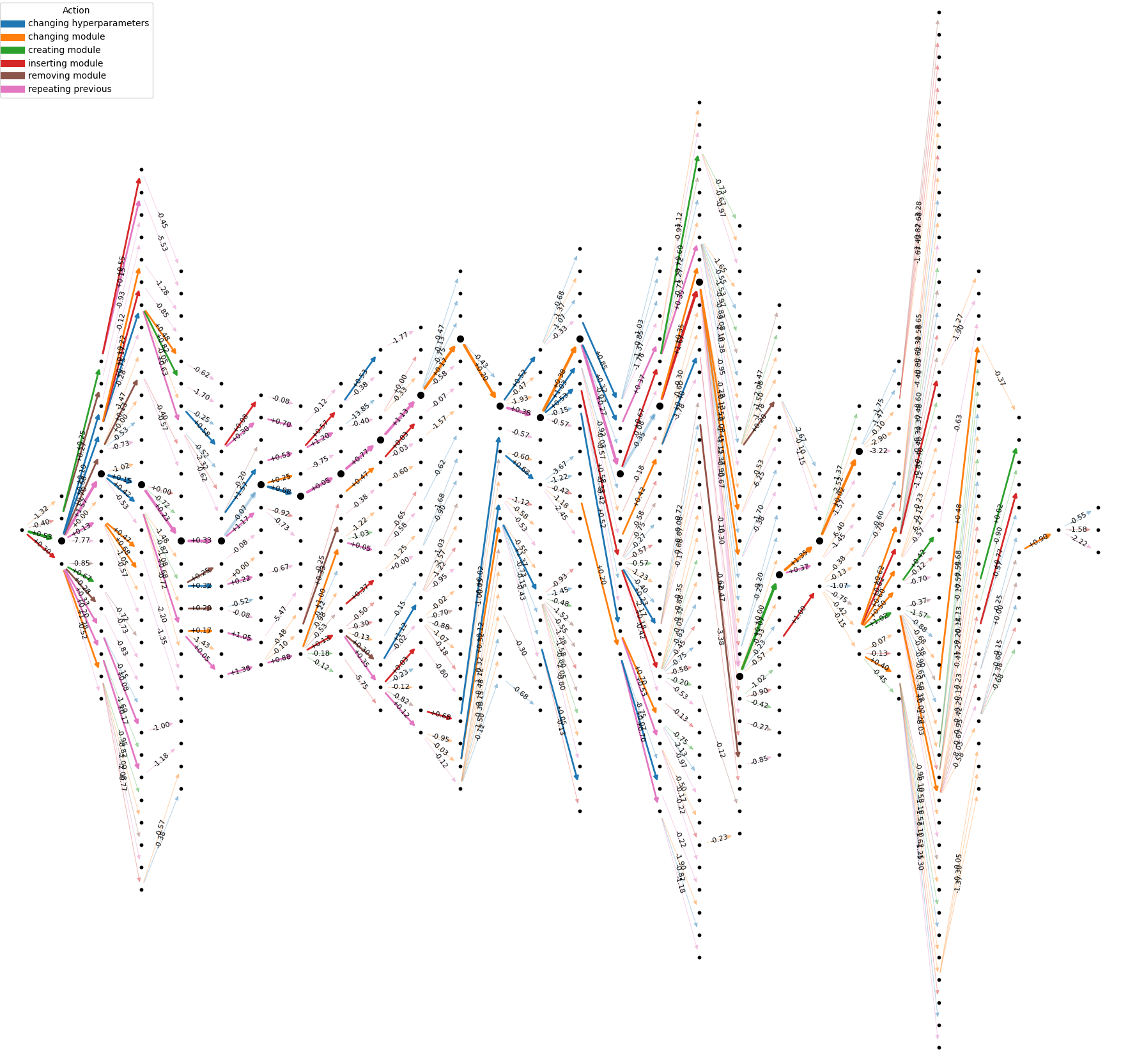}
    \vspace{-20pt}
    \caption{The evolutionary lineage of 500 architectures explored on CIFAR-100 dataset. The thick lines and large circles indicate the path that leads to the best-performing architecture. The numbers indicate the performance improvement in percentage points. The thick and thin lines indicate transformations that improved and degraded performance, respectively.}
    \label{fig:app_lineage}
\end{figure}

Fig.~\ref{fig:app_lineage} shows the evolutionary lineage of 500 architectures explored on CIFAR-100 dataset. The thick lines and large circles indicate the path that leads to the best-performing architecture. The lineage shows that performance is improved through a combination of multiple operations. It also reveals that, once a promising direction of evolution is identified, the \textit{Repeat Previous} operation is applied multiple times to repeatedly perform transformations in the same direction, further improving performance.

\subsection{Relying LLM or Minimum LLM Use}
\begin{table*}[h]
\centering
\caption{We investigate how much should we rely LLMs in \textit{swap module} operation. We conduct an investigation whether the two decisions—where to replace and which module to use—should be handled by random sampling or by the LLM’s decision. In this experiment, we only use the \textit{swap module} operation in 100 architecture evolution.
}
\label{tab:app_swap}
\scalebox{1.0}{
\begin{tabular}{ccc}
 \toprule
where to swap & which module to use & Acc. \\
\midrule
LLM &  sample 8 $\rightarrow$ LLM & 29.98 \\
sample 5 $\rightarrow$ LLM &  sample 8 $\rightarrow$ LLM & 30.14 \\
random sample & sample 8 $\rightarrow$ LLM & 34.34 \\
random sample & random sample & \textbf{35.58} \\
\bottomrule
\end{tabular}}
\end{table*}
We investigate how much we should rely on LLMs in \textit{swap module} operation. To swap a module, it is necessary to determine where to replace, which module to use, and what hyperparameters to use. Since hyperparameters can only be determined by the LLM, we conduct an investigation to decide whether the other two decisions—where to replace and which module to use—should be handled by random sampling or by the LLM’s decision.

Interestingly, Table \ref{tab:app_swap} shows that performance improves more when random sampling is used instead of relying on prior knowledge of LLMs. Of course, in practical settings where the swap module operation is not executed 100 times as in this experiment, the number of times the swap module operation is used becomes smaller, which makes it more necessary to leverage prior knowledge to improve efficiency. However, if NAS relies too heavily on prior knowledge of LLMs , the improvement tends to stagnate during the search process.

Furthermore, the results in Table \ref{tab:app_llm_size} indicate that this trend remains unchanged even when a larger LLM is used.

\subsection{Detailed Ablation Studies about Hierarchical Thompson Sampling}
\begin{table*}[ht]
\centering
\caption{Ablation studies of the exploration-exploitation trade-off in our hierarchical Thompson sampling.
}
\label{tab:app_thompson}
\scalebox{1.0}{
\begin{tabular}{cccc|c}
 \toprule
 $\epsilon$-greedy Thompson &uniform
&hierarchical& $\epsilon$&Acc.\\
 \midrule
  &$p_1$,$p_2$, $p_3$, $p_4$& -  & -& 38.86\\
 $p_1$ &$p_2$, $p_3$, $p_4$
& -& 0.5& 40.21\\
  $p_1$,$p_2$,$p_3$&$p_4$& w/o& 0.5&36.21\\
  $p_1$,$p_2$,$p_3$&$p_4$& w/& 0.5&36.38\\
 $p_1$,$p_2$&$p_3$, $p_4$& w/o& 0.5&36.88\\
  $p_1$,$p_2$&$p_3$, $p_4$& w/& 0.0&35.10\\
 $p_1$,$p_2$&$p_3$, $p_4$& w/& 0.2&34.72\\
 $p_1$,$p_2$&$p_3$, $p_4$& w/& 0.5&\textbf{40.78}\\
 $p_1$,$p_2$&$p_3$, $p_4$& w/& 0.8&39.97\\
 $p_1$,$p_2$&$p_3$, $p_4$& w/& 1.0&38.86\\
\bottomrule
\end{tabular}}
\end{table*}
In this section, we conduct a detailed ablation study of \textbf{hierarchical Thompson sampling}, which combines Bayesian sampling with coarse-to-fine hierarchical decision making.
In the proposed method, $\epsilon$-greedy Thompson sampling \citep{do2024epsilon} is applied hierarchically to $p_1$ and $p_2$ in Equation (4) defined in the main text (the decision of the operation and the prompt category).
We compare this with several alternatives: applying $\epsilon$-greedy Thompson sampling up to $p_3$, removing the hierarchical structure, and varying the value of $\epsilon$, which controls the probability of using uniform sampling instead of Thompson sampling.

As shown in Table \ref{tab:app_thompson}, when the decision process is not hierarchical, the meaningful history cannot be accumulated within 100 architecture trials, resulting in a degraded performance.
Even with hierarchical decision making, applying Bayesian exploitation up to $p_3$ makes the history too sparse because the number of effective trials is only around 60--70 after excluding the \textit{repeat previous} operation.
Therefore, a meaningful history can only be obtained when the number of combinations is limited to approximately $\text{operation}~(5) \times \text{prompt category}~(3) \approx 15$.

We also find that $\epsilon \approx 0.5$ yields the best performance.
Because the optimal strategy changes as the architecture evolves during the search, exploration should be emphasized more than exploitation through pure Thompson sampling.

\subsection{Evaluation of Each Operation}
\begin{table*}[ht]
\centering
\caption{The performance of each operation on ImageNet-100 one epoch setting.
}
\label{tab:app_operation}
\scalebox{0.9}{
\begin{tabular}{l|c}
 \toprule
 & Acc.\\
 \midrule
 MobileNet-V3& 24.21\\
  \midrule
 Change hyperparameter& 31.48\\
 Swap module& 32.94\\
 Insert module and Delete module& 31.86\\
 Create module& 31.04\\
 \midrule
 All basic operations& 34.82\\
  ~~~+ Repeat previous& 40.78\\
   \bottomrule
\end{tabular}}
\end{table*}
The performance of each operation is reported in Table \ref{tab:app_operation}. In this experiment, we evaluate each operation independently by conducting searches that use only a single type of operation. As shown in the results, each operation can improve performance. However, relying on a single operation is insufficient; combining all operations to explore diverse directions leads to substantially better performance.

Furthermore, because the search is open-ended, the \textit{repeat previous operation} significantly accelerates the exploration process. Once a direction that improves performance is discovered, repeating the same directional changes allows the search to efficiently exploit that improvement trajectory.

\subsection{Sensitivity to the Performance of LLM}
\begin{table*}[t]
\centering
\caption{Analysis of the impact of LLM performance differences. We report the results of searching 100 architectures on ImageNet-100 with one epoch training.
}
\label{tab:app_llm_size}
\scalebox{1.0}{
\begin{tabular}{cccc}
 \toprule
 & Qwen3-4B & Qwen3-8B & Qwen3-32B\\
  \midrule
LLMasTool w/o minimum LLM category & 35.52 & 36.75& 34.52\\ 
LLMasTool & \textbf{41.26} & 40.78& 39.97\\ 
\bottomrule
\end{tabular}}
\end{table*}

Table \ref{tab:app_llm_size} presents the results obtained using LLMs of different sizes. As can be seen, the proposed method achieves strong performance regardless of the capability of the LLM. This is likely because the evolution step is defined as a tree transformation, which is easier than code generation, allowing the method to function effectively even when the LLM is not highly capable. 

It may be specific to these three models, but in our experiments the smaller LLMs show slightly better performance. Since our experiments consistently improve by prioritizing diversity, one possible explanation is that stronger LLMs memorize the information in the training data more faithfully, resulting in a stronger prior bias that may hinder performance in this setting.

Furthermore, in the coarse-to-fine decision making process, prompt categories that minimally rely on LLMs are effective even when using high-performance LLMs. This suggests that, regardless of the LLM’s capability, the prior bias is largely determined by the data on which the LLM was trained. Therefore, to achieve performance beyond the existing models, it is important to conduct a search process that does not rely solely on the LLM but instead uses diversifying algorithmic exploration to avoid being constrained by the LLM’s prior bias.

\subsection{Sensitivity to the Base Architectures}

\begin{table*}[ht]
\centering
\caption{Investigation of the robustness of the proposed method with respect to the base architecture. Architectures are searched on ImageNet-100 to maximize performance after 1 epoch of training, under the constraints of 300M FLOPs and fewer than 2M parameters. 100 architectures are searched.
}
\label{tab:app_base_arch}
\scalebox{0.9}{
\begin{tabular}{l|ccc}
 \toprule
 Base arch.& Params [M] & MFLOPs & Acc.\\
 \midrule
 ShuffleNet \cite{zhang2018shufflenet} & 1.25 & 158 & 23.86 \\
 ~~+ LLMasTool & 1.52 & 186 & \textbf{39.22} \\
 \midrule
 MobileNetV3-Small0.5 \cite{koonce2021mobilenetv3} & 0.97 & 57 & 24.21 \\
 ~~+ LLMasTool & 0.95 & 202 & \textbf{40.78} \\
 \midrule
 MobileViT-XXS \cite{mehta2021mobilevit} & 0.98 & 254 & 22.00 \\
 ~~+ LLMasTool & 0.90& 239& \textbf{31.17}\\
 \bottomrule
\end{tabular}}
\end{table*}
We investigate the robustness of the proposed method with respect to the base models used as the initial data for the architecture database $\mathcal{D}_A$. In addition to MobileNetV3-Small0.5, we consider three base models of comparable scale: ShuffleNet \citep{zhang2018shufflenet}, an earlier lightweight architecture, and MobileViT-XXS \citep{mehta2021mobilevit}, which incorporates attention mechanisms. 
We search for 100 architectures on ImageNet-100 \citep{tian2020contrastive}, aiming to maximize performance after one epoch of training under the constraints of 300M FLOPs and fewer than 2M parameters.

As shown in Table \ref{tab:app_base_arch}, the performance can be substantially improved from all base models. However, when the search is limited to only 100 explored architectures, the results are still affected by the choice of the base model, which can lead to lower performance. In particular, when training for only one epoch, the MobileViT-based models perform poorly because attention modules tend to converge slowly. Nevertheless, in the best discovered architecture, all attention modules are successfully removed. This indicates that the method is not constrained by the prior assumption that attention mechanisms are inherently more advanced than CNNs, and instead can evolve architectures that are better suited to the target task setting.

\subsection{ImageNet-100 with Longer Epochs}
\label{sec:app_imagenet100_100ep_exp}
\begin{table*}[ht]
\centering
\caption{Ablation Studies of the proposed components in the architecture evolution. This time, we maximize the accuracy with 100 epoch training on ImageNet-100.  We report results after generating 100 architectures. The grayed components are our proposal. 
}
\label{tab:ablation100e}
\scalebox{0.9}{
\begin{tabular}{ccc|c}
 \toprule
 evo. step &planning &sampling&Acc.\\
 \midrule
 MobileNetV3  & -  & - & 73.00
\\
 code-gen & LLM's idea & uniform& 74.96\\
 \cellcolor{gray!30} tree-trans  & \cellcolor{gray!30} c-to-f & uniform&81.00\\
 \cellcolor{gray!30} tree-trans  & \cellcolor{gray!30} c-to-f & \cellcolor{gray!30} H-Thompson&\textbf{81.72}\\
 \bottomrule
\end{tabular}}
\end{table*}
In the ablation study in the main paper, we searched for models that achieve high accuracy after only one epoch of training. Demonstrating large performance improvements under this setting suggests that LLMasTool is effective even in practical scenarios where training must be performed with limited data, as well as in settings that deviate from commonly used training protocols.

On the other hand, it is also important to examine whether, under more conventional training settings that align well with the information contained in LLM training data, it is still preferable to use the LLM as a tool through coarse-to-fine decision making, rather than delegating all decisions to the LLM in an agentic manner. To investigate this, Table \ref{tab:ablation100e} also presents an ablation study where 100 architectures are searched on ImageNet-100 with 100 training epochs. Detailed experimental settings are described in Sec. \ref{sec:app_imagenet_setup}.

As shown in the table, coarse-to-fine decision making remains important even under standard training settings, indicating that exploration that is not constrained by the prior biases of the LLM is crucial. Furthermore, hierarchical Thompson sampling is also shown to be effective.

% \subsection{Searching for ImageNet-1K}

\section{Implementation Details}
\subsection{Overall Evolution Flow}
An overview of the evolution process described in the main text is shown in Algorithm \ref{alg:evolution} . The general framework of the evolution process follows a standard evolutionary procedure.
\begin{algorithm}[t]
\caption{Architecture Evolution}
\label{alg:evolution}
\small
\begin{algorithmic}[1]
\Require Module database $\mathcal{D}_M$, initial architecture database $\mathcal{D}_A$
\State Initialize history database $\mathcal{D}_H \leftarrow \emptyset$

\While{search budget not exhausted}

\State Sample base architecture
\[
A_t^{base} \sim \mathcal{D}_A
\]

\State Sample tree transformation using Algorithm~\ref{alg:cf_sampling}
\[
T_t \sim p_{cf}\!\left(T \mid A_t^{base}, \mathcal{D}_A, \mathcal{D}_M, \mathcal{D}_H , \mathcal{LLM}\right)
\]

\State Apply transformation
\[
A_t \leftarrow T_t(A_t^{base})
\]

\State Check feasibility (executability, size constraint, intended transformation)
\[
\epsilon_t = \mathcal{E}(A_t^{base}, A_t, T_t)
\]

\If{$\epsilon_t = 1$}
    \State Instantiate $A_t$ and train/evaluate to obtain a metric $e_t$
    
    \State Update architecture database
    \[
    \mathcal{D}_A \leftarrow \mathcal{D}_A \cup \{(A_t, e_t)\}
    \]

    \State Update history database
    \[
    \mathcal{D}_H \leftarrow \mathcal{D}_H \cup \{(T_t, A_t^{base}, e_t - e_t^{base})\}
    \]
\EndIf

\EndWhile
\end{algorithmic}
\end{algorithm}

Algorithm \ref{alg:evolution} presents the proposed coarse-to-fine decision-making algorithm. Because the architecture is represented in a tree structure that is easy for the algorithm to manipulate, the algorithm can explicitly determine the direction of architectural modifications. We introduce a gradation in how much responsibility is assigned to the LLM. Some prompt categories allow the LLM to decide many of the decisions, while others restrict the LLM to making only minimal decisions. The details are described in the next section.
\begin{algorithm}[t]
\caption{Coarse-to-Fine Decision making $p_{cf}$}
\label{alg:cf_sampling}
\small
\begin{algorithmic}[1]
\Require Base architecture $A_t^{base}$, module database $\mathcal{D}_M$, history database $\mathcal{D}_H$

\State Sample operation $o \sim p_1(o)$ via $\epsilon$-greedy Thompson sampling using $\mathcal{D}_H$

\State Sample prompt category conditioned on operation $c \sim p_2(c \mid o)$ via $\epsilon$-greedy Thompson sampling using $\mathcal{D}_H$

\State Sample prompt template $\tau \sim p_3(\tau \mid c,o)$ uniformly

\State Retrieve candidate nodes/modules from $A_t^{base}$ and $\mathcal{D}_M$

\State Sample placeholder assignments $\phi \sim p_4(\phi \mid c,o,\tau,A_t^{base},\mathcal{D}_M)$

\State Construct prompt using $(o,c,\tau,\phi)$ and history $\mathcal{D}_H$

\State Use LLM to generate concrete tree transformation \\
\hspace{1.5em} $T \sim p_5(T \mid c,o,\tau,\phi,A_t^{base},\mathcal{D}_M,\mathcal{D}_H)$

\State \Return $T$
\end{algorithmic}
\end{algorithm}

\subsection{Tree Transformation Operations}
\label{sec:app_operation_details}
The proposed tree transformation $T$ is realized through a set of tools that operate on the architecture tree.
These tools consist of (i) algorithm-based tools that deterministically analyze and manipulate the tree structure, and (ii) LLM-based tools that enable transformations beyond the capability of predefined algorithms.

When processed by algorithmic tools, an architecture tree $A$ is represented as a nested Python dictionary object that may contain list structures.
Formally, we denote this representation as
\begin{equation}
A \in \mathcal{T}_{\text{dict}},
\end{equation}
where $\mathcal{T}_{\text{dict}}$ denotes the space of nested dictionary-based tree structures.
In contrast, when interacting with the LLM, the architecture is converted into a textual configuration format similar to the OpenMMLab configuration file, denoted as
\begin{equation}
A_{\text{text}} \in \mathcal{T}_{\text{text}}.
\end{equation}

We first introduce the algorithm-based tools used to analyze and manipulate the tree structure.

\begin{itemize}

\item \textbf{Module Attribute Extraction}  
A function that retrieves all module attributes and their corresponding module names appearing in the tree structure:
\begin{equation}
\mathrm{Attr}(A) = \{ (a_1, m_1), (a_2, m_2), \ldots , (a_K, m_K) \},
\end{equation}
where $a_k$ denotes the hierarchical attribute of the module \\(e.g., \textit{model.backbone.module1}) and $m_k$ denotes the corresponding module name.

\item \textbf{Sequential Module Attribute Extraction}  
A function that extracts module attributes located inside the list structures of the tree:
\begin{equation}
\mathrm{Attr}_{\mathrm{list}}(A) \subseteq \mathrm{Attr}(A).
\end{equation}
In most architectures, list structures correspond to sequential module compositions.

\item \textbf{Sub-tree Retrieval}  
A function that retrieves the sub-tree (module) located at a specified attribute $a$ from the architecture $A$:
\begin{equation}
S = \mathrm{GetSubTree}(A, a),
\end{equation}
where $a$ denotes the hierarchical attribute of a module in $A$, and $S$ denotes the corresponding sub-tree.

\item \textbf{Sub-tree Replacement}  
A function that replaces a sub-tree (module) located at a specified attribute $a$ with a new sub-tree $S$:
\begin{equation}
A' = \mathrm{Replace}(A, a, S).
\end{equation}

\item \textbf{Sub-tree Deletion from List Structure}  
A function that removes the sub-tree (module) corresponding to a specified attribute $a$ within a list structure:
\begin{equation}
A' = \mathrm{Delete}_{\mathrm{list}}(A, a).
\end{equation}

\item \textbf{Sub-tree Insertion into List Structure}  
A function that inserts a new sub-tree (module) $S$ immediately after the attribute $a$ within a list structure:
\begin{equation}
A' = \mathrm{Insert}_{\mathrm{list}}(A, a, S).
\end{equation}

\item \textbf{Module Source Code Retrieval}  
A function that retrieves the source code of specified one or several modules from the module database $\mathcal{D}_M$:
\begin{equation}
C = \mathrm{GetCode}(\{m\}, \mathcal{D}_M),
\end{equation}
where $m$ denotes the module name and $C$ denotes the corresponding source code retrieved from the database.

\item \textbf{Default Parameter Retrieval}  
A function that retrieves the default initialization parameters of a specified module from the module database $\mathcal{D}_M$:
\begin{equation}
\theta = \mathrm{GetDefault}(m, \mathcal{D}_M),
\end{equation}
where $m$ denotes the module name and $\theta$ denotes the set of default initialization parameters associated with the module.

\item \textbf{Tree-to-Text Conversion}  
A function that converts a nested Python dictionary object into a textual configuration format:
\begin{equation}
A_{text} = \mathrm{Text}(A).
\end{equation}

\item \textbf{Text-to-Tree Conversion}  
A function that converts the textual configuration format back into a nested Python dictionary object:
\begin{equation}
A = \mathrm{Dict}(A_{text}).
\end{equation}

\item \textbf{Module Retrieval from Database}  
This function takes one or more modules as input and retrieves candidate modules from the module database $\mathcal{D}_M$ whose forward interfaces are compatible with the input modules. Specifically, the retrieved modules must have a forward function whose numbers of input and output variables can match those of at least one input module.

\begin{equation}
\mathcal{M} = \mathrm{Retrieve}(\mathcal{S}, \mathcal{D}_M),
\end{equation}

where $\mathcal{S}=\{m_1,\ldots,m_k\}$ denotes the set of input module names and $\mathcal{M}\subseteq\mathcal{D}_M$ denotes the set of retrieved candidate module names.

\end{itemize}

The following operations are executed by using the LLM as a tool in combination with the above algorithm-based tools.

\subsubsection{Change Hyperparameter Operation.}
As shown in the prompt of Sec. \ref{sec:prompt_param}, hyperparameter modifications are not handled primarily through dedicated tools; instead, the algorithm only provides coarse guidance, and the LLM directly outputs the resulting architecture. This choice is because changing only one hyperparameter is not efficient.

\subsubsection{Swap Module Operation.}

We swap modules as shown in Algorithm \ref{alg:swap} using the above algorithm-based tools in conjunction with the LLM.
\begin{algorithm}[h!]
\caption{Swap Module Operation}
\label{alg:swap}
\small
\begin{algorithmic}[1]

\Require Base architecture $A_t^{base}$, module database $\mathcal{D}_M$, history database $\mathcal{D}_H$, prompt category $c$, prompt template $P$

\State $\mathcal{A} = \{(a_1,m_1),\ldots,(a_K,m_K)\} \leftarrow \mathrm{Attr}(A_t^{base})$
\Comment Extract all module attributes

\If{$c \in \{\text{rely LLM},\ \text{involve LLM}\}$}
    \If{$P \in \mathcal{P}_{restrict}$}
        \State Randomly sample a subset $\tilde{\mathcal{A}} \subset \mathcal{A}$ and replace $\mathcal{A} \leftarrow \tilde{\mathcal{A}}$
        \Comment We sampled 5
    \EndIf
    \State $\mathcal{M} \leftarrow \mathrm{Retrieve}(\{m_1,\ldots,m_K\}, \mathcal{D}_M)$
    \Comment Candidate module names
    \State Randomly sample candidate modules $\tilde{\mathcal{M}} \subset \mathcal{M}$
    \Comment We sampled 8
    \State $P' = P.\mathrm{format}(\mathrm{Text}(A_t^{base}), \{a_1,\ldots,a_K\}, \mathrm{GetCode}(\tilde{\mathcal{M}}, \mathcal{D}_M), \mathcal{D}_H)$\\
    \Comment Fill blanks in the prompt template
    \State $(a_d, m') = \mathcal{LLM}(P')$
    \Comment $a_d$: swap location, $m'$: new module
\Else
    \Comment minimum LLM
    \State Randomly sample $(a_d,m_d)$ from $\mathcal{A}$
    \State $\mathcal{M} \leftarrow \mathrm{Retrieve}(\{m_d\}, \mathcal{D}_M)$
    \If{$P$ requires LLM}
        \State Randomly sample candidate modules $\tilde{\mathcal{M}} \subset \mathcal{M}$
        \State $P' = P.\mathrm{format}(\mathrm{Text}(A_t^{base}), a_d, \mathrm{GetCode}(\tilde{\mathcal{M}}, \mathcal{D}_M), \mathcal{D}_H)$\\
        \Comment Fill blanks in the prompt template
        \State $m' = \mathcal{LLM}(P')$
        \Comment Decide $m'$ from $\tilde{\mathcal{M}}$
    \Else
        \State $m'$ is randomly sampled from $\mathcal{M}$
    \EndIf
\EndIf

\State $u = \mathrm{Text}(\mathrm{GetSubTree}(A_t^{base}, a_d))$
\Comment Get parameters used in the previous module

\State Let $P_2$ be a prompt template to determine the hyperparameters of $m'$

\State $P'_2 = P_2.\mathrm{format}(\mathrm{GetCode}(\{m_d,m'\}, \mathcal{D}_M), u, \mathrm{GetDefault}(m', \mathcal{D}_M))$

\State $S' = \mathrm{Dict}(\mathcal{LLM}(P'_2))$
\Comment Generated sub-tree with the new module

\State $A_t = \mathrm{Replace}(A_t^{base}, a_d, S')$
\Comment Replace the module at $a_d$ with $S'$

\State \Return transformed architecture $A_t$

\end{algorithmic}
\end{algorithm}

\subsubsection{Insert Module Operation.}

We insert modules as shown in Algorithm \ref{alg:insert}.

\begin{algorithm}[h!]
\caption{Insert Module Operation}
\label{alg:insert}
\small
\begin{algorithmic}[1]

\Require Base architecture $A_t^{base}$, module database $\mathcal{D}_M$, history database $\mathcal{D}_H$, prompt category $c$, prompt template $P$

\State $\mathcal{A} = \{(a_1,m_1),\ldots,(a_K,m_K)\} \leftarrow \mathrm{Attr_{list}}(A_t^{base})$\\
\Comment Extract all module attributes in list structures

\If{$c \in \{\text{rely LLM},\ \text{involve LLM}\}$}
    \State Randomly sample a subset $\tilde{\mathcal{A}} \subset \mathcal{A}$ and replace $\mathcal{A} \leftarrow \tilde{\mathcal{A}}$ 
    \Comment We sampled 5
    \State $\mathcal{M} \leftarrow \mathrm{Retrieve}(\{m_1,\ldots,m_K\}, \mathcal{D}_M)$ 
    \Comment Candidate module names
    \State Randomly sample candidate modules $\tilde{\mathcal{M}} \subset \mathcal{M}$ 
    \Comment We sampled 8
    \State $P' = P.\mathrm{format}(\mathrm{Text}(A_t^{base}), \{a_1,\ldots,a_K\}, \mathrm{GetCode}(\tilde{\mathcal{M}}, \mathcal{D}_M), \mathcal{D}_H)$\\
    \Comment Fill blanks in the prompt template
    \State $(a_d, m') = \mathcal{LLM}(P')$
    \Comment $a_d$: insertion location, $m'$: new module
\Else
    \Comment minimum LLM
    \State Randomly sample $(a_d,m_d)$ from $\mathcal{A}$
    \State $\mathcal{M} \leftarrow \mathrm{Retrieve}(\{m_d\}, \mathcal{D}_M)$
    \If{$P$ requires LLM}
        \State Randomly sample candidate modules $\tilde{\mathcal{M}} \subset \mathcal{M}$
        \State $P' = P.\mathrm{format}(\mathrm{Text}(A_t^{base}), a_d, \mathrm{GetCode}(\tilde{\mathcal{M}}, \mathcal{D}_M), \mathcal{D}_H)$
        \Comment Fill blanks in the prompt template
        \State $m' = \mathcal{LLM}(P')$
        \Comment Decide $m'$ from $\tilde{\mathcal{M}}$
    \Else
        \State $m'$ is randomly sampled from $\mathcal{M}$
    \EndIf
\EndIf

\State $u = \mathrm{Text}(\mathrm{GetSubTree}(A_t^{base}, a_d))$
\Comment Get parameters used in the previous module

\State Let $P_2$ be a prompt template to determine the hyperparameters of $m'$

\State $P'_2 = P_2.\mathrm{format}(\mathrm{GetCode}(\{m_d,m'\}, \mathcal{D}_M), u, \mathrm{GetDefault}(m', \mathcal{D}_M))$

\State $S' = \mathrm{Dict}(\mathcal{LLM}(P'_2))$
\Comment Generated sub-tree with the new module

\State $A_t = \mathrm{Insert_{list}}(A_t^{base}, a_d, S')$
\Comment Insert $S'$ just after attribute $a_d$

\State \Return transformed architecture $A_t$

\end{algorithmic}
\end{algorithm}

\subsubsection{Remove Module Operation.}

We remove modules as shown in Algorithm \ref{alg:remove}.

\begin{algorithm}[h!]
\caption{Remove Module Operation}
\label{alg:remove}
\small
\begin{algorithmic}[1]

\Require Base architecture $A_t^{base}$, module database $\mathcal{D}_M$, history database $\mathcal{D}_H$, prompt category $c$, prompt template $P$

\State $\mathcal{A}=\{(a_1,m_1),\ldots,(a_K,m_K)\}\leftarrow \mathrm{AttrList}(A_t^{base})$
\Comment Extract all module attributes and corresponding module names

\If{$c \in \{\text{rely LLM},\ \text{inv.\ LLM}\}$}
    \State Randomly sample a subset $\tilde{\mathcal{A}} \subset \mathcal{A}$
    \Comment We sampled 5 candidates
    \State $\mathcal{A} \leftarrow \tilde{\mathcal{A}}$
    
    \State $P' = P.\mathrm{format}\big(
        \mathrm{Text}(A_t^{base}),
        \{a_1,\ldots,a_K\},
        \mathrm{GetCode}(\{m_1,\ldots,m_K\},\mathcal{D}_M),
        \mathcal{D}_H
    \big)$\\
    \Comment Fill blanks in the prompt template
    
    \State $a_d = \mathcal{LLM}(P')$
    \Comment Attribute to remove, chosen from $\{a_1,\ldots,a_K\}$

\Else \Comment minimum LLM
    \State Randomly sample $(a_d,m_d)$ from $\mathcal{A}$
\EndIf

\State $u = \mathrm{Text}(\mathrm{GetSubTree}(A_t^{base}, a_d))$
\Comment Used parameters of the module to be removed

\State Let $P_2$ be a prompt template to generate the full arch. after removing the module

\State $P'_2 = P_2.\mathrm{format}\big(
    \mathrm{Text}(A_t^{base}),
    u,
    \mathrm{GetCode}(\{m_d\},\mathcal{D}_M)
\big)$

\State $A_t = \mathrm{Dict}(\mathcal{LLM}(P'_2))$
\Comment Generate the architecture after removal

\State \Return transformed architecture $A_t$

\end{algorithmic}
\end{algorithm}

\subsubsection{Create Module Operation.}

We create composite modules as shown in Algorithm \ref{alg:create}.

\begin{algorithm}[h!]
\caption{Create Module Operation}
\label{alg:create}
\scriptsize
\begin{algorithmic}[1]

\Require Base architecture $A_t^{base}$, module database $\mathcal{D}_M$, history database $\mathcal{D}_H$, prompt category $c$, prompt template $P$

\State Let primitive modules and special modules be $\mathcal{M}_{pytorch}$ and $\mathcal{M}_{special}$

\State $b_{merge} \leftarrow (P \in \mathcal{P}_{merge})$
\Comment Whether the prompt requires merging multiple modules

\If{$b_{merge}$} \Comment{Merge several modules into a composite module}

    \State $\mathcal{A} \leftarrow \mathrm{Attr_{list}}(A_t^{base})$
    \Comment Extract module attributes in list structures

    \State Randomly sample $(a_d,m_d)$ from $\mathcal{A}$

    \State Randomly sample the number of merging modules $n$

    \If{$P \in \mathcal{P}_{restrict}$}
        \State $\mathcal{M}_{candidate} \leftarrow \{m_d,\ldots,m_{d+n}\} \cup \mathcal{M}_{special}$
    \Else
        \State Sample custom modules $\mathcal{M}_{custom}$ from $\mathcal{D}_M$
        \State $\mathcal{M}_{candidate} \leftarrow \{m_d,\ldots,m_{d+n}\} \cup \mathcal{M}_{special} \cup \mathcal{M}_{custom}$
    \EndIf

\Else \Comment{Replace a module with a composite module}

    \State $\mathcal{A} \leftarrow \mathrm{Attr}(A_t^{base})$
    \Comment Extract all module attributes

    \State Randomly sample $(a_d,m_d)$ from $\mathcal{A}$

    \If{$P \in \mathcal{P}_{restrict}$}
        \State $\mathcal{M}_{candidate} \leftarrow \{m_d\} \cup \mathcal{M}_{special}$
    \Else
        \State Sample custom modules $\mathcal{M}_{custom}$ from $\mathcal{D}_M$
        \State $\mathcal{M}_{candidate} \leftarrow \{m_d\} \cup \mathcal{M}_{special} \cup \mathcal{M}_{custom}$
    \EndIf

\EndIf

\If{$c \in \{\text{inv. LLM},\ \text{minimum LLM}\}$}
    \State Randomly sample $\tilde{\mathcal{M}}_{pytorch} \subset \mathcal{M}_{pytorch}$
    \State $\mathcal{M}_{pytorch} \leftarrow \tilde{\mathcal{M}}_{pytorch}$
\EndIf

\If{$c \in \{\text{rely LLM},\ \text{inv. LLM}\}$}
    \State $C' \leftarrow \text{None}$ \Comment No constraint
\Else
    \State Let $C$ be a constraint template
    \State Sample $m_s \sim \mathcal{M}_{special}$
    \State $C' \leftarrow C.\text{format}(m_d, m_s)$ \Comment A constraint to use at least $m_d$ and $m_s$
\EndIf

% \State $P' \leftarrow P.\text{format}(
%     \mathrm{Text}(A_t^{base}),
%     \mathrm{GetCode}(\mathcal{M}_{candidate},\mathcal{D}_M),
%     \mathrm{GetDefault}(\mathcal{M}_{special},\mathcal{D}_M),
%     \mathrm{GetSubTree}(A_t^{base}, a_d),
%     \mathcal{D}_H,
%     C'
% )$
\State $\begin{aligned}
P' \leftarrow P.\text{format}(&\mathrm{Text}(A_t^{base}), \mathrm{GetCode}(\mathcal{M}_{candidate}, \mathcal{D}_M), \mathrm{GetDefault}(\mathcal{M}_{special}, \mathcal{D}_M), \\
&\mathrm{GetSubTree}(A_t^{base}, a_d), \mathcal{D}_H, C')
\end{aligned}$

\State $S_{text} \leftarrow \mathcal{LLM}(P')$ \Comment Ask LLM to create a new module using $\mathcal{M}_{candidate}$ and $\mathcal{M}_{special}$ with constraint $C'$\\

\State $S' \leftarrow \mathrm{Dict}(S_{text})$
\Comment Convert LLM output into a sub-tree

\State $A_t \leftarrow \mathrm{Replace}(A_t^{base}, a_d, S')$

\If{$b_{merge}$}
    \For{$i = n$ \textbf{down to} $1$}
        \State $A_t \leftarrow \mathrm{Remove}_{list}(A_t, a_{d+i})$
    \EndFor
\EndIf

\State \Return $A_t$

\end{algorithmic}
\end{algorithm}

\subsection{History Database}

The history database is used both to improve efficiency through Thompson sampling in coarse-to-fine decision making and to provide contextual information to the LLM. Here, we describe the details of the textualized history that is passed to the LLM.

First, the operations \textit{"Swap Module," "Insert Module,"} and \textit{"Remove Module"} explicitly track the exact transformation that has been performed. Therefore, we generate accurate history entries using rule-based templates in the following formats:

\begin{itemize}
    \item \texttt{"Change \{module\_name\_pre\} at \{attribute\} into \{module\_name\_new\}"}
    \item \texttt{"Insert \{module\_name\_new\} at \{attribute\}"}
    \item \texttt{"Remove \{module\_name\_pre\} at \{attribute\}"}
\end{itemize}

On the other hand, the operations \textit{"Change Hyperparameter," "Create Module,"} and \textit{"Repeat Previous"} may involve diverse and complex modifications, making them difficult to handle with rule-based algorithms. Therefore, after the transformation is completed, we prompt the LLM with the instruction: \textit{“Please describe the transformation in short one sentence.”} The LLM then generates the textualized history entry.

When presenting the history to the LLM, the entries are shown in a format illustrated in the \textit{Instruction Prompt (Turn 1)} in Sec. \ref{sec:prompt_overall}.

\subsubsection{Repeat Previous Operation.}
As shown in the prompt of Sec. \ref{sec:prompt_repeat}, hyperparameter modifications are not handled primarily through dedicated tools; instead, the algorithm only provides coarse guidance, and the LLM directly outputs the resulting architecture. This choice is because it is difficult for rule-based algorithm to extract what is the similar directional change to previous transformation.

\subsection{Prompts}
\label{sec:prompt}
\subsubsection{Overall Pipeline.}
\label{sec:prompt_overall}
In our framework, the final tree transformation is determined through an algorithm-guided instruction process combined with a multi-turn interaction with the LLM.

First, the LLM is provided with the following \textit{system prompt}. 

\begin{tcolorbox}[
    enhanced, 
    colback=yellow!5!white, 
    colframe=yellow!75!black, 
    title=System Prompt
] 
You are a helpful assistant to create a Python-based config file of the MMPretrain. Your mission is to create an innovative model architecture which outperforms previous works. Take account the information you will be provided and create a great model config. Don't talk about dataloader, augmentation, optimizer, and learning rate. You just need to create the model config. You are not allowed to use pretrained weight.
\end{tcolorbox}

In the first turn, we additionally input the \textbf{Instruction Prompt (Turn 1)} shown below. In this prompt, one of the prompt templates described in the following sections is used based on which operation to use. Based on the proposed coarse-to-fine decision making, the algorithm selects which prompt category and prompt template to use. The placeholders in the selected template are then filled using algorithm-based analysis of the tree structure, and the resulting prompt is provided to the LLM.

Afterward, depending on the operation, the algorithm and the LLM interact for several turns, and the final transformation is determined through this dialogue process.

In the next section, we present examples of the prompt templates and multi-turn conversations for each operation.

\begin{tcolorbox}[
    enhanced, 
    breakable,
    colback=yellow!5!white, 
    colframe=yellow!75!black, 
    title=Instruction Prompt (Turn 1),
] 
% \textcolor{Green}{\# Following three prompts are randomly used.}\\
% ------------------------------------------------------------------------------------------ \\
\scriptsize
Please improve the config: \textcolor{blue}{\{pre\_cfg\}}\\
\\
Below is a summary of previous experiments. Analyze what could improve performance and consider what to do next.\\
\\
------\\
\# Previous experiments\\
\textcolor{blue}{\{HISTORY\}}
\\
\\
Tips for utilizing previous experiments:\\
- Reverting to previous modules/hyperparameters is not interesting because we already know the performance of the previous model. Do not do that!\\
- Changing the same positions repeatedly is inefficient. Extract the essence and apply it to other positions.\\
- We want an innovative model, so trying new things that have never been done is important.\\
------\\
\\
\textcolor{blue}{\{Operation Specific Prompt Template\}}\\
\\

% ------------------------------------------------------------------------------------------ \\
% Can you create a more innovative model config than \textcolor{blue}{\{pre\_cfg\}}?\\
% \textcolor{blue}{\{Operation Specific Prompt Template\}}\\
% ------------------------------------------------------------------------------------------ \\
% Please improve the performance of the config: \textcolor{blue}{\{pre\_cfg\}}\\
% \textcolor{blue}{\{Operation Specific Prompt Template\}}\\
% \\

\textcolor{Green}{
------------------------------------------------------------------------------------------ \\
\# Example of \{pre\_cfg\}\\
------------------------------------------------------------------------------------------ \\
```python\\
model = dict(
\begin{list}{}{\leftmargin=1.5em \topsep=0pt}
\item type=’ImageClassifier’,
\item backbone = ... \# See Appendix \ref{sec:app_discovered} for an example.
\item neck = ...
\item head = ...
\end{list}
)\\
```\\
------------------------------------------------------------------------------------------ \\
\# Example of \{HISTORY\}\\
------------------------------------------------------------------------------------------ \\
------\\
Changes: Remove BasicBlock at model.backbone.layer\_cfgs[4]
Performance: Deteriorated\\
------\\
Changes: Added a third ConvFFN at layer\_cfgs[31] with 256 internal channels and 0.4 drop path for enhanced feature refinement.\\  
Performance: Improved\\
------\\
...\\
------\\
------------------------------------------------------------------------------------------ \\
}
\end{tcolorbox}

\subsubsection{Change Hyperparameters.}
\label{sec:prompt_param}

We prepare the following prompt templates for \textit{Change Hyperparameters} operation. One of these templates is used as the last part of the \textbf{Instruction Prompt (Turn 1)}. In this operation, the algorithm does not explicitly decide which specific parameters to modify because changing only a single parameter at a time is inefficient. The algorithm narrows down a set of candidate locations and instructs the LLM to modify several hyperparameters at once. The LLM then generates the entire tree architecture accordingly.

\begin{tcolorbox}[
    enhanced, 
    breakable,
    colback=yellow!5!white, 
    colframe=yellow!75!black, 
    title=Prompt Templates for Change Hyperparameters,
] 
\scriptsize
\textcolor{Green}{\# Category: Rely LLM.\\
Four templates with following variations.\\
\textcolor{orange}{\{{GUIDANCE}\}} = [
\begin{list}{}{\leftmargin=1.5em \topsep=0pt}
\item "changing a few hyperparameters in the config",
\item "changing some important hyperparameters in the config",
\item "changing hyperparameters that are likely to affect model performance",
\item "modifying hyperparameters which you think weird in the config"
\end{list}
]\\
}
\textcolor{Green}{\# Category: Inv. LLM.\\
Three templates with following variations.\\
\textcolor{orange}{\{{GUIDANCE}\}} = [
\begin{list}{}{\leftmargin=1.5em \topsep=0pt}
\item "changing hyperparameters that have not been changed until now",
\item "changing hyperparameters to values not tried before",
\item "changing hyperparameters to unexpected values"
\end{list}
]\\
}
\textcolor{Green}{\# Category: Mini. LLM.\\
14 templates with following variations.\\
\textcolor{orange}{\{{GUIDANCE}\}} = [
\begin{list}{}{\leftmargin=1.5em \topsep=0pt}
\item "slightly increasing the channel width of some layers",
\item "slightly decreasing the channel width of some layers",
\item "slightly increasing the channel width of the initial layers",
\item "slightly decreasing the channel width of the initial layers",
\item "slightly increasing the channel width of the middle layers",
\item "slightly decreasing the channel width of the middle layers",
\item "slightly increasing the channel width of the final layers",
\item "slightly decreasing the channel width of the final layers",
\item "changing some boolean hyperparameters in the config",
\item "changing some float hyperparameters in the config",
\item "changing some integer hyperparameters in the config",
\item "changing some string hyperparameters in the config",
\item "changing the receptive field of some layers",
\item "changing the stride of some layers"
\end{list}
]\\
}
    
------------------------------------------------------------------------------------------ \\
This time, please improve the model's performance by \textcolor{orange}{\{{GUIDANCE}\}}. Change only hyperparameters. Do not add additional hyperparameters. Just change the values of existing hyperparameters. Also, do not change the architecture or modules. In other words, do not change the value of each "type". Please provide a fully modified model config.
------------------------------------------------------------------------------------------ \\
\\

\end{tcolorbox}

\subsubsection{Swap Module.}

We prepare the following prompt templates for \textit{Swap Module} operation. One of these templates is used as the last part of the \textbf{Instruction Prompt (Turn 1)}. As the prompt indicates, in the first turn of the conversation, the LLM determines which part should be replaced with which module if the algorithm has not decided yet.

\begin{tcolorbox}[
    enhanced, 
    breakable,
    colback=yellow!5!white, 
    colframe=yellow!75!black, 
    title=Prompt Templates for Swap Module,
] 
\scriptsize
\textcolor{Green}{\# Category: Rely LLM.\\
2 $\times$ 2 = 4 templates with following variations.\\
\textcolor{orange}{\{{GUIDANCE}\}} = [
\begin{list}{}{\leftmargin=1.5em \topsep=0pt}
\item "Positions with Some Issues",
\item "Positions Previously Not Changed Much"
\end{list}
]\\
}
\textcolor{orange}{\{CANDIDATE\_ATTRIBUTES\}} \textcolor{Green}{= [}
\begin{list}{}{\leftmargin=1.5em \topsep=0pt}
\item \textcolor{blue}{\{all\_module\_attributes\}},
\item \textcolor{blue}{\{sampled\_module\_attributes\}}
\end{list}
\textcolor{Green}{]
}

------------------------------------------------------------------------------------------ \\
This time, let's try to improve the model's performance by using one of the following modules:\\
\textcolor{blue}{\{candidate\_module\_codes\}}\\
Analyze the source codes, especially how they work and what the input and output tensor shapes are, to replace previous modules. Then, which module should be used and where should it be replaced in the original config? \\
I will ask later about appropriate hyperparameters, so please answer in the following format.\\
\textcolor{Green}{\#\#\#\#\#\#\#\#\#\#}\\
Knowledge from Previous Experiments: XXX\\
\textcolor{orange}{\{{GUIDANCE}\}} (More Improvement Needed Positions): XXX\\
Pros and Cons of the New Candidate Modules: XXX\\
Input and Output Shape Compatibility of the New Candidate Modules against Previous Modules: XXX\\
New Module Name to Use: YYY\\
Where to be Used: ZZZ\\
\textcolor{Green}{\#\#\#\#\#\#\#\#\#\#}\\
Fill XXX based on your analysis.\\
YYY can be chosen from \textcolor{blue}{\{candidate\_module\_names\}}.\\
Select ZZZ from \textcolor{orange}{\{CANDIDATE\_ATTRIBUTES\}}.\\
------------------------------------------------------------------------------------------ \\
\\

\textcolor{Green}{\# Category: Inv. LLM.\\
Two templates.}\\
------------------------------------------------------------------------------------------ \\
This time, let's try to improve the model's performance by using one of the following modules:\\
\textcolor{blue}{\{candidate\_module\_codes\}},\\
but I want to try an unexpected module this time.\
Analyze the source codes, especially how they work and what the input and output tensor shapes are, to replace previous modules.\\
Then, which module should be used and where should it be replaced in the original config? 
I will ask later about appropriate hyperparameters, so please answer in the following format.\\
\textcolor{Green}{\#\#\#\#\#\#\#\#\#\#}\\
Knowledge from Previous Experiments: XXX\\
Positions Previously Not Changed Much (More Improvement Needed Positions): XXX\\
\textbf{New Candidate Modules that few researchers would consider (This time, I want to try novel changes)}: XXX\\
Input and Output Shape Compatibility of the New Candidate Modules against Previous Modules: XXX\\
New Module Name to Use: YYY\\
Where to be Used: ZZZ\\
\textcolor{Green}{\#\#\#\#\#\#\#\#\#\#}\\
Fill XXX based on your analysis.\\
YYY can be chosen from \textcolor{blue}{\{candidate\_module\_names\}}.\\
Select ZZZ from \textcolor{blue}{\{sampled\_module\_attributes\}}.\\
------------------------------------------------------------------------------------------ \\
This time, let's try to improve the model's performance by using one of the following modules:\\
\textcolor{blue}{\{candidate\_module\_codes\}}\\
Analyze the source codes, especially how they work and what the input and output tensor shapes are, to replace previous modules.\\ 
Then, which module should be used and where should it be replaced in the original config?\\ 
I will ask later about appropriate hyperparameters, so please answer in the following format.\\
\textcolor{Green}{\#\#\#\#\#\#\#\#\#\#}\\
Knowledge from Previous Experiments: XXX\\
\textbf{Positions rarely modified  (This time, I want to try novel changes)}: XXX\\
Pros and Cons of the New Candidate Modules: XXX\\
Input and Output Shape Compatibility of the New Candidate Modules against Previous Modules: XXX\\
New Module Name to Use: YYY\\
Where to be Used: ZZZ\\
\textcolor{Green}{\#\#\#\#\#\#\#\#\#\#}\\
Fill XXX based on your analysis.\\
YYY can be chosen from \textcolor{blue}{\{candidate\_module\_names\}}.\\
Select ZZZ from \textcolor{blue}{\{sampled\_module\_attributes\}}.\\
------------------------------------------------------------------------------------------ \\
\\

\textcolor{Green}{\# Category: Mini. LLM.\\
Two templates but one of them skips to ask to LLM by randomly deciding where to swap and which module to use.}\\
------------------------------------------------------------------------------------------ \\
This time, let's try to improve the model's performance by replacing the module in \{decided\_module\_attribute\} with one of the following modules:\\
\textcolor{blue}{\{candidate\_module\_codes\}}\\
Analyze the source codes, especially about how they work and what are the input and output tensor shapes to replace with the previous module.\\ 
Then, answer which module should be used.\\
I'll ask later about appropriate hyperparameters, so please answer in the following format.\\
\textcolor{Green}{\#\#\#\#\#\#\#\#\#\#}\\
Knowledge from Previous Experiments: XXX\\
Pros and Cons of the New Candidate Modules: XXX\\
Input and Output Shape Compatibility of the New Candidate Modules against Previous\\ Modules: XXX\\
New Module Name to Use: YYY\\
\textcolor{Green}{\#\#\#\#\#\#\#\#\#\#}\\
Fill XXX based on your analysis.\\
YYY can be chosen from \textcolor{blue}{\{sampled\_module\_names\}}.\\
------------------------------------------------------------------------------------------ \\
(SKIP)\\
------------------------------------------------------------------------------------------ \\
\\
\end{tcolorbox}

In the second turn of the conversation, appropriate hyperparameters of the new module are determined using the following instruction prompt. The obtained hyperparameters are then extracted, and the tree structure is actually transformed in an algorithmic manner.

\begin{tcolorbox}[
    enhanced, 
    breakable,
    colback=yellow!5!white, 
    colframe=yellow!75!black, 
    title=Instruction Prompt for Swap Module Operation (Turn 2),
] 
\scriptsize
Now, I want to ask how to replace. The original module, \textcolor{blue}{\{original\_module\_name\}}, is \\
\textcolor{blue}{\{original\_module\_code\}},\\
and the used parameters of \_\_init\_\_ function are \\
\textcolor{blue}{\{used\_parameters\}}. \\
The new module, \textcolor{blue}{\{decided\_module\_name\}}, is\\
\textcolor{blue}{\{decided\_module\_code\}}, \\
and the default parameters of \_\_init\_\_ function are \\
\textcolor{blue}{\{decided\_module\_default\_param\}}.\\
To replace the original module with the new module, modifications to the default parameters are needed to ensure the input and output tensor shapes match those of the original module.  \\
If there is a "<TODO>" in the default parameters, it must be replaced with an appropriate parameter. Moreover, further modification of the default parameters might be desired to improve the performance. However, adding new parameters (keys of the dict) are not allowed.\\
Considering above, please modify the default parameters of \\
\textcolor{blue}{\{decided\_module\_default\_param\}}. \\
Note that you need to modify the values of the dict and don't need to modify the keys of the dict.
\end{tcolorbox}

\subsubsection{Insert Module.}

We prepare the following prompt templates for \textit{Insert Module} operation. One of these templates is used as the latter part of the \textbf{Instruction Prompt (Turn 1)}. If the algorithm has not yet decided, the LLM determines which module should be inserted and where it should be placed.

\begin{tcolorbox}[
    enhanced, 
    breakable,
    colback=yellow!5!white, 
    colframe=yellow!75!black, 
    title=Prompt Templates for Insert Module,
] 
\scriptsize
\textcolor{Green}{\# Category: Rely LLM.\\
Two templates.}\\

------------------------------------------------------------------------------------------ \\
This time, let's try to improve the model's performance by inserting an additional module after one of the following positions: \textcolor{blue}{\{sampled\_module\_attributes\}}. We can use one of the following modules:\\
\textcolor{blue}{\{candidate\_module\_codes\}}\\
Analyze the source codes, especially about how they work and what are the input and output tensor shapes. Then, answer which module should be used and where it should be inserted. I'll ask later about appropriate hyperparameters, so please answer in the following format.\\
\#\#\#\#\#\#\#\#\#\#\\
Knowledge from Previous Experiments: XXX\\
Positions Previously Not Inserted Much (More Improvement Might be Possible): XXX\\
Pros and Cons of the Candidate Modules: XXX\\
Analysis of the Candidate Inserting Positions: XXX\\
Input and Output Shape Compatibility of the Candidate Modules to Insert: XXX\\
New Module Name to Use: YYY\\
Where to be Inserted: ZZZ\\
\#\#\#\#\#\#\#\#\#\#\\
Fill XXX based on your analysis.\\
YYY can be chosen from \textcolor{blue}{\{candidate\_module\_names\}}.\\
ZZZ can be chosen from \textcolor{blue}{\{sampled\_module\_attributes\}}.\\
------------------------------------------------------------------------------------------ \\
This time, let's try to improve the model's performance by inserting an additional module after one of the following positions: \textcolor{blue}{\{sampled\_module\_attributes\}}. We can use one of the following modules:\\
\textcolor{blue}{\{candidate\_module\_codes\}}\\
Analyze the source codes, especially about how they work and what are the input and output tensor shapes. Then, answer which module should be used and where it should be inserted. I'll ask later about appropriate hyperparameters, so please answer in the following format.\\
\#\#\#\#\#\#\#\#\#\#\\
Knowledge from Previous Experiments: XXX\\
Pros and Cons of the Candidate Modules: XXX\\
Positions Previously Not Inserted Much (More Improvement Might be Possible): XXX\\
Analysis about Where Lacks the Expressiveness: XXX\\
Input and Output Shape Compatibility of the Candidate Modules to Insert: XXX\\
New Module Name to Use: YYY\\
Where to be Inserted: ZZZ\\
\#\#\#\#\#\#\#\#\#\#\\
Fill XXX based on your analysis.\\
YYY can be chosen from \textcolor{blue}{\{candidate\_module\_names\}}.\\
ZZZ can be chosen from \textcolor{blue}{\{sampled\_module\_attributes\}}.\\
------------------------------------------------------------------------------------------ \\
\\

\textcolor{Green}{\# Category: Inv. LLM.\\
Two templates.}\\
------------------------------------------------------------------------------------------ \\
This time, let's try to improve the model's performance by inserting an additional module after one of the following positions: \textcolor{blue}{\{sampled\_module\_attributes\}}. We can use one of the following modules:\\
\textcolor{blue}{\{candidate\_module\_codes\}},\\
but I want to try an unexpected module this time.\\
Analyze the source codes, especially about how they work and what are the input and output tensor shapes. Then, answer which module should be used and where it should be inserted. I'll ask later about appropriate hyperparameters, so please answer in the following format.\\
\#\#\#\#\#\#\#\#\#\#\\
Knowledge from Previous Experiments: XXX\\
Positions Previously Not Inserted Much (More Improvement Might be Possible): XXX\\
New Candidate Modules that few researchers would consider (This time, I want to try novel changes): XXX\\
Analysis of the Candidate Inserting Positions: XXX\\
Input and Output Shape Compatibility of the Candidate Modules to Insert: XXX\\
New Module Name to Use: YYY\\
Where to be Inserted: ZZZ\\
\#\#\#\#\#\#\#\#\#\#\\
Fill XXX based on your analysis.\\
YYY can be chosen from \textcolor{blue}{\{candidate\_module\_names\}}.\\
ZZZ can be chosen from \textcolor{blue}{\{sampled\_module\_attributes\}}.\\
------------------------------------------------------------------------------------------ \\
This time, let's try to improve the model's performance by inserting an additional module after one of the following positions: \textcolor{blue}{\{sampled\_module\_attributes\}}, especially focusing on novel changes. We can use one of the following modules:\\
\textcolor{blue}{\{candidate\_module\_codes\}}\\
Analyze the source codes, especially about how they work and what are the input and output tensor shapes. Then, answer which module should be used and where it should be inserted. I'll ask later about appropriate hyperparameters, so please answer in the following format.\\
\#\#\#\#\#\#\#\#\#\#\\
Knowledge from Previous Experiments: XXX\\
Pros and Cons of the Candidate Modules: XXX\\
Positions where Nobody will Insert (This time, I want to try novel changes): XXX\\
Input and Output Shape Compatibility of the Candidate Modules to Insert: XXX\\
New Module Name to Use: YYY\\
Where to be Inserted: ZZZ\\
\#\#\#\#\#\#\#\#\#\#\\
Fill XXX based on your analysis.\\
YYY can be chosen from \textcolor{blue}{\{candidate\_module\_names\}}.\\
ZZZ can be chosen from \textcolor{blue}{\{sampled\_module\_attributes\}}.\\
------------------------------------------------------------------------------------------ \\
\\

\textcolor{Green}{\# Category: Mini. LLM.\\
Two templates but one of them skips to ask to LLM by randomly deciding where to replace and which module to use.}\\
------------------------------------------------------------------------------------------ \\
This time, let's try to improve the model's performance by inserting an additional module after \textcolor{blue}{\{decided\_module\_attribute\}}. We can use one of the following modules:\\
\textcolor{blue}{\{candidate\_module\_codes\}}\\
Analyze the source codes, especially about how they work and what are the input and output tensor shapes to insert after \textcolor{blue}{\{decided\_module\_attribute\}}. Then, answer which module should be used. I'll ask later about appropriate hyperparameters, so please answer in the following format.\\
\textcolor{Green}{\#\#\#\#\#\#\#\#\#\#}\\
Knowledge from Previous Experiments: XXX\\
Pros and Cons of the Candidate Modules: XXX\\
Input and Output Shape Compatibility of the Candidate Modules to Insert After \textcolor{blue}{\{decided\_module\_attribute\}}: XXX\\
New Module Name to Use: YYY\\
\textcolor{Green}{\#\#\#\#\#\#\#\#\#\#}\\
Fill XXX based on your analysis.\\
YYY can be chosen from \textcolor{blue}{\{candidate\_module\_names\}}.\\
------------------------------------------------------------------------------------------ \\
(SKIP)\\
------------------------------------------------------------------------------------------ \\
\\
\end{tcolorbox}

In the second turn of the conversation, appropriate hyperparameters of the new module are determined using the following instruction prompt. The obtained hyperparameters are then extracted, and the tree structure is actually transformed in an algorithmic manner.

\begin{tcolorbox}[
    enhanced, 
    breakable,
    colback=yellow!5!white, 
    colframe=yellow!75!black, 
    title=Instruction Prompt for Insert Module Operation (Turn 2),
] 
\scriptsize
Now, I want to ask how to insert it. The previous module, \textcolor{blue}{\{original\_module\_name\}}, is \\
\textcolor{blue}{\{original\_module\_code\}},\\
and the used parameters of \_\_init\_\_ function are \\
\textcolor{blue}{\{used\_parameters\}}. \\
The new module, \textcolor{blue}{\{decided\_module\_name\}}, is\\
\textcolor{blue}{\{decided\_module\_code\}}, \\
and the default parameters of \_\_init\_\_ function are \\
\textcolor{blue}{\{decided\_module\_default\_param\}}.\\
To insert it, modifications to the default parameters are needed to ensure that the input shape of the new module matches the output shape of the preceding module and that its output shape is compatible with the subsequent module. \\
If there is a "<TODO>" in the default parameters, it must be replaced with an appropriate parameter. Moreover, further modification of the default parameters might be desired to improve the performance. However, adding new parameters (keys of the dict) are not allowed.\\
Considering above, please modify the default parameters of \\
\textcolor{blue}{\{decided\_module\_default\_param\}}. \\
Note that you need to modify the values of the dict and don't need to modify the keys of the dict.
\end{tcolorbox}

\subsubsection{Remove Module.}

We prepare the following prompt templates for \textit{Remove Module} operation. One of these templates is used as the latter part of the \textbf{Instruction Prompt (Turn 1)}. If the algorithm has not yet decided, the LLM determines which module should be removed.

\begin{tcolorbox}[
    enhanced, 
    breakable,
    colback=yellow!5!white, 
    colframe=yellow!75!black, 
    title=Prompt Templates for Remove Module,
] 
\scriptsize
\textcolor{Green}{\# Category: Rely LLM.\\
Two templates.}\\
------------------------------------------------------------------------------------------ \\
This time, let's try to improve the model's performance by removing a module from \textcolor{blue}{\{sampled\_module\_attributes\}}. The source codes of the modules are as follows:\\
\textcolor{blue}{\{candidate\_module\_codes\}}\\
Analyze the source codes. Then, answer which module should be removed. I'll ask later about appropriate hyperparameters, so please answer in the following format.\\
\#\#\#\#\#\#\#\#\#\#\\
Knowledge from Previous Experiments: XXX\\
Positions Previously Not Removed Much (More Improvement Might be Possible): XXX\\
Positions Where You Think Weird: XXX\\
Where to be Removed: YYY\\
\#\#\#\#\#\#\#\#\#\#\\
Fill XXX based on your analysis.\\
YYY can be chosen from \textcolor{blue}{\{sampled\_module\_attributes\}}.\\
------------------------------------------------------------------------------------------ \\
This time, let's try to improve the model's efficiency by removing a module from \textcolor{blue}{\{sampled\_module\_attributes\}}. The source codes of the modules are as follows:\\
\textcolor{blue}{\{candidate\_module\_codes\}}\\
Analyze the source codes. Then, answer which module should be removed. I'll ask later about appropriate hyperparameters, so please answer in the following format.\\
\#\#\#\#\#\#\#\#\#\#\\
Knowledge from Previous Experiments: XXX\\
Positions Previously Not Removed Much (More Improvement Might be Possible): XXX\\
Where to be Removed: YYY\\
\#\#\#\#\#\#\#\#\#\#\\
Fill XXX based on your analysis.\\
YYY can be chosen from \textcolor{blue}{\{sampled\_module\_attributes\}}.\\
------------------------------------------------------------------------------------------ \\
\\

\textcolor{Green}{\# Category: Inv. LLM.\\
Two templates.}\\
------------------------------------------------------------------------------------------ \\
This time, let's try to improve the model's performance by removing a module from \textcolor{blue}{\{sampled\_module\_attributes\}}, especially focusing on unexpected positions. The source codes of the modules are as follows:\\
\textcolor{blue}{\{candidate\_module\_codes\}}\\
Analyze the source codes. Then, answer which module should be removed. I'll ask later about appropriate hyperparameters, so please answer in the following format.\\
\#\#\#\#\#\#\#\#\#\#\\
Knowledge from Previous Experiments: XXX\\
Rarely considered removal locations (This time, I want to try novel changes): XXX\\
Where to be Removed: YYY\\
\#\#\#\#\#\#\#\#\#\#\\
Fill XXX based on your analysis.\\
YYY can be chosen from \textcolor{blue}{\{sampled\_module\_attributes\}}.\\
------------------------------------------------------------------------------------------ \\
This time, let's try to improve the model's efficiency by removing a module from \textcolor{blue}{\{sampled\_module\_attributes\}}, especially focusing on novel changes. The source codes of the modules are as follows:\\
\textcolor{blue}{\{candidate\_module\_codes\}}\\
Analyze the source codes. Then, answer which module should be removed. I'll ask later about appropriate hyperparameters, so please answer in the following format.\\
\#\#\#\#\#\#\#\#\#\#\\
Knowledge from Previous Experiments: XXX\\
Positions where You think Important (This time, I want to try novel changes): XXX\\
Where to be Removed: YYY\\
\#\#\#\#\#\#\#\#\#\#\\
Fill XXX based on your analysis.\\
YYY can be chosen from \textcolor{blue}{\{sampled\_module\_attributes\}}.\\
------------------------------------------------------------------------------------------ \\
\\

\textcolor{Green}{\# Category: Mini. LLM.\\
One template which skips to ask to LLM by randomly deciding where to remove.}\\
------------------------------------------------------------------------------------------ \\
(SKIP)\\
------------------------------------------------------------------------------------------ \\
\\
\end{tcolorbox}

In the second turn of the conversation, the complete configuration after removing the decided module is requested. Although we request full architecture generation, the following verification process check whether the intended module is removed.

\begin{tcolorbox}[
    enhanced, 
    breakable,
    colback=yellow!5!white, 
    colframe=yellow!75!black, 
    title=Instruction Prompt for Remove Module Operation (Turn 2),
] 
\scriptsize
Now, I want to ask how to remove the module at \textcolor{blue}{\{decided\_module\_attribute\}}. I want you to create a config without this module. You might need to modify hyperparameters for surrounding modules at \textcolor{blue}{\{surrounding\_module\_attributes\}} to make the model work well without the removed module.\\
First, refer to the following source codes of relevant modules:\\
\textcolor{blue}{\{decided\_module\_code\}}\\
\textcolor{blue}{\{surrounding\_modules\_code\}}\\
Then, please provide the complete model config after removing the module at \textcolor{blue}{\{decided\_module\_attribute\}} and modifying hyperparameters of surrounding modules as needed.
\end{tcolorbox}

\subsubsection{Create Module.}

We prepare the following prompt templates for \textit{Create Module} operation. One of these templates is used as the latter part of the \textbf{Instruction Prompt (Turn 1)}. In this operation, a composite module is created to replace one module or several modules. This operation obtains the final output within a single turn of conversation, and the algorithm replaces the corresponding part of the tree structure with the composite module.

\begin{tcolorbox}[
    enhanced, 
    breakable,
    colback=yellow!5!white, 
    colframe=yellow!75!black, 
    title=Prompt Templates for Create Module,
] 
\scriptsize
\textcolor{Green}{\# Category: Rely LLM.\\
2 $\times$ 2 = 4 templates with following variations.}\\
\textcolor{orange}{\{{HOW\_AND\_WHERE}\}} \textcolor{Green}{= [
\begin{list}{}{\leftmargin=1.5em \topsep=0pt}
\item "replacing \textcolor{blue}{\{decided\_module\_attribute\}} with",
\item "merging \textcolor{blue}{\{num\}} modules at \textcolor{blue}{\{decided\_sequential\_attributes\}} into"
\end{list}
]\\
}
\textcolor{orange}{\{CUSTOM\_MODULES\}} \textcolor{Green}{= [}
\begin{list}{}{\leftmargin=1.5em \topsep=0pt}
\item \textcolor{Green}{Original module:}\textcolor{blue}{\{original\_module\_code\}},
\item \textcolor{Green}{Original modules and Custom modules:}\textcolor{blue}{\{original\_module\_code\}} + \textcolor{blue}{\{custom\_modules\_code\}}
\end{list}
\textcolor{Green}{]\\
}
------------------------------------------------------------------------------------------ \\
This time, let's try to improve the model's performance by \textcolor{orange}{\{{HOW\_AND\_WHERE}\}} a new module created by combining some of the following modules:\\
PyTorch modules (I only show the \_\_init\_\_ parameters since you know well):\textcolor{blue}{\{pytorch\_modules\_dict\}}\\
\textcolor{orange}{\{CUSTOM\_MODULES\}}\\
Special modules which flexibly combine these modules with config dictionaries:\textcolor{blue}{\{special\_modules\_code\}}\\
For example, you can create a module like as follows:\\
{
```python\\
dict(type='ParallelWithConfig',
    \begin{list}{}{\leftmargin=1.5em \topsep=0pt}
    \item module\_cfg1=dict(
        \begin{list}{}{\leftmargin=1.5em \topsep=0pt}
        \item type='SequentialWithConfig',
        \item module\_cfgs=[
            \begin{list}{}{\leftmargin=1.5em \topsep=0pt}
            \item dict(type='Conv2d', in\_channels=32, out\_channels=64, 
                \begin{list}{}{\leftmargin=1.5em \topsep=0pt}
                \item kernel\_size=1, stride=1, padding=0, dilation=1, groups=1, 
                \item bias=True, padding\_mode='zeros', device=None, dtype=None),
            \end{list}
            \item dict(type='GELU', approximate='tanh'),
            \item dict(type='Conv2d', in\_channels=64, out\_channels=32, 
                \begin{list}{}{\leftmargin=1.5em \topsep=0pt}
                \item kernel\_size=3, stride=1, padding=1, dilation=1, groups=1, 
                \item bias=True, padding\_mode='zeros', device=None, dtype=None),
                \end{list}
            \item dict(type='Sigmoid')
            \end{list}
        \item ]
        \end{list}
    \item ),
    \item module\_cfg2=dict(type='Identity'),
    \item merge\_operation='mul',
    \end{list}
)\\
```\\
}
Note that we have to make the input and output tensor shapes compatible with the previous module(s) with the following parameters:\textcolor{blue}{\{used\_parameters\}}\\
Please answer in the following format.\\
\#\#\#\#\#\#\#\#\#\#\\
Knowledge from Previous Experiments: XXX\\
Input and Output Shape of the Previous Module(s): XXX\\
New Module Configuration:\\
```python\\
XXX\\
```\\
\#\#\#\#\#\#\#\#\#\#\\
Fill XXX based on your analysis.\\
------------------------------------------------------------------------------------------ \\
\\

\textcolor{Green}{\# Category: Inv. LLM.\\
2 $\times$ 2 = 4 templates with following variations.}\\
\textcolor{orange}{\{{HOW\_AND\_WHERE}\}} \textcolor{Green}{= [
\begin{list}{}{\leftmargin=1.5em \topsep=0pt}
\item "replacing \textcolor{blue}{\{decided\_module\_attribute\}} with",
\item "merging \textcolor{blue}{\{num\}} modules at \textcolor{blue}{\{decided\_sequential\_attributes\}} into"
\end{list}
]\\
}
\textcolor{orange}{\{CUSTOM\_MODULES\}} \textcolor{Green}{= [}
\begin{list}{}{\leftmargin=1.5em \topsep=0pt}
\item \textcolor{Green}{Original module:}\textcolor{blue}{\{original\_module\_code\}},
\item \textcolor{Green}{Original modules and Custom modules:}\textcolor{blue}{\{original\_module\_code\}} + \textcolor{blue}{\{custom\_modules\_code\}}
\end{list}
\textcolor{Green}{]\\
}
------------------------------------------------------------------------------------------ \\
This time, let's try to improve the model's performance by \textcolor{orange}{\{{HOW\_AND\_WHERE}\}} a new module created by combining some of the following modules:\\
PyTorch modules (I only show the \_\_init\_\_ parameters since you know well):\textcolor{blue}{\{pytorch\_modules\_dict\}}\\
\textcolor{orange}{\{CUSTOM\_MODULES\}}\\
Special modules which flexibly combine these modules with config dictionaries:\textcolor{blue}{\{special\_modules\_code\}}\\
For example, you can create a module like as follows:\\
{
```python\\
dict(type='ParallelWithConfig',
    \begin{list}{}{\leftmargin=1.5em \topsep=0pt}
    \item module\_cfg1=dict(
        \begin{list}{}{\leftmargin=1.5em \topsep=0pt}
        \item type='SequentialWithConfig',
        \item module\_cfgs=[
            \begin{list}{}{\leftmargin=1.5em \topsep=0pt}
            \item dict(type='Conv2d', in\_channels=32, out\_channels=64, 
                \begin{list}{}{\leftmargin=1.5em \topsep=0pt}
                \item kernel\_size=1, stride=1, padding=0, dilation=1, groups=1, 
                \item bias=True, padding\_mode='zeros', device=None, dtype=None),
            \end{list}
            \item dict(type='GELU', approximate='tanh'),
            \item dict(type='Conv2d', in\_channels=64, out\_channels=32, 
                \begin{list}{}{\leftmargin=1.5em \topsep=0pt}
                \item kernel\_size=3, stride=1, padding=1, dilation=1, groups=1, 
                \item bias=True, padding\_mode='zeros', device=None, dtype=None),
                \end{list}
            \item dict(type='Sigmoid')
            \end{list}
        \item ]
        \end{list}
    \item ),
    \item module\_cfg2=dict(type='Identity'),
    \item merge\_operation='mul',
    \end{list}
)\\
```\\
}
Note that we have to make the input and output tensor shapes compatible with the previous module(s) with the following parameters:\textcolor{blue}{\{used\_parameters\}}\\
\textbf{I expect you to create a novel module configuration that nobody has tried before, so please try to combine the modules in a unique way.
}\\
Please answer in the following format.\\
\#\#\#\#\#\#\#\#\#\#\\
Knowledge from Previous Experiments: XXX\\
Input and Output Shape of the Previous Module: XXX\\
New Module Configuration:\\
```python\\
XXX\\
```\\
\#\#\#\#\#\#\#\#\#\#\\
Fill XXX based on your analysis.\\
------------------------------------------------------------------------------------------ \\
\\

\textcolor{Green}{\# Category: Mini. LLM.\\
2 $\times$ 2 = 4 templates with following variations.}\\
\textcolor{orange}{\{{HOW\_AND\_WHERE}\}} \textcolor{Green}{= [
\begin{list}{}{\leftmargin=1.5em \topsep=0pt}
\item "replacing \textcolor{blue}{\{decided\_module\_attribute\}} with",
\item "merging \textcolor{blue}{\{num\}} modules at \textcolor{blue}{\{decided\_sequential\_attributes\}} into"
\end{list}
]\\
}
\textcolor{orange}{\{CUSTOM\_MODULES\}} \textcolor{Green}{= [}
\begin{list}{}{\leftmargin=1.5em \topsep=0pt}
\item \textcolor{Green}{Original module:}\textcolor{blue}{\{original\_module\_code\}},
\item \textcolor{Green}{Original modules and Custom modules:}\textcolor{blue}{\{original\_module\_code\}} + \textcolor{blue}{\{custom\_modules\_code\}}
\end{list}
\textcolor{Green}{]\\
}
------------------------------------------------------------------------------------------ \\
This time, let's try to improve the model's performance by \textcolor{orange}{\{{HOW\_AND\_WHERE}\}} a new module created by combining some of the following modules:\\
PyTorch modules (I only show the \_\_init\_\_ parameters since you know well):\textcolor{blue}{\{pytorch\_modules\_dict\}}\\
\textcolor{orange}{\{CUSTOM\_MODULES\}}\\
Special modules which flexibly combine these modules with config dictionaries:\textcolor{blue}{\{special\_modules\_code\}}\\
For example, you can create a module like as follows:\\
{
```python\\
dict(type='ParallelWithConfig',
    \begin{list}{}{\leftmargin=1.5em \topsep=0pt}
    \item module\_cfg1=dict(
        \begin{list}{}{\leftmargin=1.5em \topsep=0pt}
        \item type='SequentialWithConfig',
        \item module\_cfgs=[
            \begin{list}{}{\leftmargin=1.5em \topsep=0pt}
            \item dict(type='Conv2d', in\_channels=32, out\_channels=64, 
                \begin{list}{}{\leftmargin=1.5em \topsep=0pt}
                \item kernel\_size=1, stride=1, padding=0, dilation=1, groups=1, 
                \item bias=True, padding\_mode='zeros', device=None, dtype=None),
            \end{list}
            \item dict(type='GELU', approximate='tanh'),
            \item dict(type='Conv2d', in\_channels=64, out\_channels=32, 
                \begin{list}{}{\leftmargin=1.5em \topsep=0pt}
                \item kernel\_size=3, stride=1, padding=1, dilation=1, groups=1, 
                \item bias=True, padding\_mode='zeros', device=None, dtype=None),
                \end{list}
            \item dict(type='Sigmoid')
            \end{list}
        \item ]
        \end{list}
    \item ),
    \item module\_cfg2=dict(type='Identity'),
    \item merge\_operation='mul',
    \end{list}
)\\
```\\
}
Note that we have to make the input and output tensor shapes compatible with the previous module(s) with the following parameters:\textcolor{blue}{\{used\_parameters\}}\\
\textbf{This time, please use the \textcolor{blue}{\{original\_module\_name\}} and \textcolor{blue}{\{random\_special\_module\_name\}} at least.}\\
\textbf{Creating a previously unexplored module is important.}\\
Please answer in the following format.\\
\#\#\#\#\#\#\#\#\#\#\\
Knowledge from Previous Experiments: XXX\\
Input and Output Shape of the Previous Module: XXX\\
New Module Configuration:\\
```python\\
XXX\\
```\\
\#\#\#\#\#\#\#\#\#\#\\
Fill XXX based on your analysis.\\
------------------------------------------------------------------------------------------ \\
\end{tcolorbox}

\subsubsection{Repeat Previous.}
\label{sec:prompt_repeat}

We prepare the following prompt for \textit{Repeat Previous} operation.
Note that the prompt, \textbf{Instruction Prompt (Turn 1)}, is not used. Instead, the following prompts are used to directly generate the modified architecture. Because the history is sparse, we uniformly sample from the predefined prompt variations.

\begin{tcolorbox}[
    enhanced, 
    breakable,
    colback=yellow!5!white, 
    colframe=yellow!75!black, 
    title=Prompt for Repeat Previous,
] 
\scriptsize
\textcolor{Green}{\# For all previous operations}\\
\textcolor{orange}{\{{INTRODUCTION}\}} \textcolor{Green}{= [
\begin{list}{}{\leftmargin=1.5em \topsep=0pt}
\item "From the previous config:  \textcolor{blue}{\{pre\_pre\_cfg\}} we created a following config:  \textcolor{blue}{\{pre\_cfg\}} by "\textcolor{blue}{\{pre\_transform\}}". Then, the accuracy was improved from \textcolor{blue}{\{pre\_pre\_acc\}}\% to \textcolor{blue}{\{pre\_acc\}}\%.",
\item "The previous config was  \textcolor{blue}{\{pre\_pre\_cfg\}}. By "\textcolor{blue}{\{pre\_transform\}}", we successfully achieved better accuracy with the following config:  \textcolor{blue}{\{pre\_cfg\}}.",
\item "The previous config was  \textcolor{blue}{\{pre\_pre\_cfg\}}. From this config, performance is improved by "\textcolor{blue}{\{pre\_transform\}}". The config is  \textcolor{blue}{\{pre\_cfg\}}.",
\item "The previous config (Accuracy: \textcolor{blue}{\{pre\_pre\_acc\}}\%) was  \textcolor{blue}{\{pre\_pre\_cfg\}}. From this config, we improved the accuracy to \textcolor{blue}{\{pre\_acc\}}\% by "\textcolor{blue}{\{pre\_transform\}}". The improved config is  \textcolor{blue}{\{pre\_cfg\}}.",
\item "The config:  \textcolor{blue}{\{pre\_cfg\}} is created from the previous config:  \textcolor{blue}{\{pre\_pre\_cfg\}} by "\textcolor{blue}{\{pre\_transform\}}", resulting in an accuracy improvement from \textcolor{blue}{\{pre\_pre\_acc\}}\% to \textcolor{blue}{\{pre\_acc\}}\%.",
\item "By "\textcolor{blue}{\{pre\_transform\}}", we successfully created a better config:  \textcolor{blue}{\{pre\_cfg\}}.",
\item "The config:  \textcolor{blue}{\{pre\_cfg\}} is created by "\textcolor{blue}{\{pre\_transform\}}"."
\end{list}
]\\
}

\textcolor{Green}{\# Restrict prompt is prepared for each operation}\\
\textcolor{Green}{\# Change Hyperparameters}\\
\textcolor{orange}{\{RESTRICTION\}} \textcolor{Green}{= [
\begin{list}{}{\leftmargin=1.5em \topsep=0pt}
\item "extracting the essence and refining it",
\item "extracting the essence and improving the new config furthermore",
\item "making the same change but with a slightly larger magnitude",
\item "slightly increasing the strength of the change",
\item "slightly increasing the strength of the change in the same positions",
\item "applying the same or similar changes to other hyperparameters in the same modules",
\item "applying the same or similar changes to other hyperparameters in the surrounding modules",
\item "applying the same or similar changes to other hyperparameters in the far away modules",
\item "applying the same or similar changes to other hyperparameters in one different module"
\end{list}
]\\
}
\textcolor{Green}{\# Swap Module}\\
\textcolor{orange}{\{RESTRICTION\}} \textcolor{Green}{= [
\begin{list}{}{\leftmargin=1.5em \topsep=0pt}
\item "extracting the essence and improving the new config furthermore",
\item "replacing one or two other positions into \textcolor{blue}{\{module\_new\}}",
\item "replacing a module around \textcolor{blue}{\{location\}} into \textcolor{blue}{\{module\_new\}}",
\item "replacing a module in the far away positions from \textcolor{blue}{\{location\}} into \textcolor{blue}{\{module\_new\}}",
\end{list}
]\\
}
\textcolor{Green}{\# Insert Module}\\
\textcolor{orange}{\{RESTRICTION\}} \textcolor{Green}{= [
\begin{list}{}{\leftmargin=1.5em \topsep=0pt}
\item "extracting the essence and improving the new config furthermore",
\item "inserting one more \textcolor{blue}{\{module\_new\}} before the \textcolor{blue}{\{location\}}",
\item "inserting one more \textcolor{blue}{\{module\_new\}} after the \textcolor{blue}{\{location\}}",
\item "inserting one more \textcolor{blue}{\{module\_new\}} in a far away position from \textcolor{blue}{\{location\}}",
\item "inserting one more \textcolor{blue}{\{module\_new\}} at \textcolor{blue}{\{random\_location\}"}
\end{list}
]\\
}
\textcolor{Green}{\# Remove Module}\\
\textcolor{orange}{\{RESTRICTION\}} \textcolor{Green}{= [
\begin{list}{}{\leftmargin=1.5em \topsep=0pt}
\item "extracting the essence and improving the new config furthermore",
\item "removing a module at \textcolor{blue}{\{location\}} from the new config",
\item "removing a module at \textcolor{blue}{\{location\}} from the new config",
\item "removing a module working similar to \textcolor{blue}{\{module\_pre\}}",
\item "removing a similar module to \textcolor{blue}{\{module\_pre\}} around \textcolor{blue}{\{location\}}",
\item "removing a similar module to \textcolor{blue}{\{module\_pre\}} far away from \textcolor{blue}{\{location\}}"
\end{list}
]\\
}
\textcolor{Green}{\# Create Module and Repeat Previous}\\
\textcolor{orange}{\{RESTRICTION\}} \textcolor{Green}{= [
\begin{list}{}{\leftmargin=1.5em \topsep=0pt}
\item "extracting the essence and improving the new config furthermore",
\item "doing something similar to the new config",
\item "making a further change in the same direction to the new config"
\end{list}
]\\
}
    
------------------------------------------------------------------------------------------ \\
\textcolor{orange}{\{INTRODUCTION\}}\\
I think the change, "\textcolor{blue}{\{pre\_transform\}}", is a good idea. So, this time, I want you to further improve the performance by \textcolor{orange}{\{RESTRICTION\}}.\\
You can refer the source code of relevant modules below: \textcolor{blue}{\{relevant\_source\_code\}}\\
Please provide the improved complete config.\\
------------------------------------------------------------------------------------------ \\
\\

\end{tcolorbox}

\section{Discovered Architectures}
\label{sec:app_discovered}
An example of the discovered architecture is shown below.
The tree representation is converted into an OpenMMLab \citep{OpenMMLab}-style configuration format, and this format is presented to the LLM as \textit{pre\_cfg} in the prompt.

\section{ImageNet16-120}

\begin{tcblisting}{boxrule=2pt, listing only,listing options={language=iPython},title={Discovered architecture for ImageNet16-120 \\(Acc. 53.20\%, 37.3 MFLOPs, 0.63 \#Params[M])}, breakable}
model = dict(
    # mmpretrain.models.classifiers.image.ImageClassifier
    type='ImageClassifier',
    backbone=dict(
        # nas_special_modules.NAS_Backbone
        type='NAS_Backbone',
        layer_cfgs=[
            dict(
                # mmcv.cnn.bricks.conv_module.ConvModule
                type='ConvModule',
                in_channels=3,
                out_channels=16,
                kernel_size=3,
                stride=1,
                padding=1,
                dilation=1,
                groups=1,
                bias=False,
                with_spectral_norm=False,
                order=(
                    'conv',
                    'norm',
                    'act',
                ),
                norm_cfg=dict(
                    # torch.nn.BatchNorm2d
                    type='BatchNorm2d', 
                    eps=1e-05, 
                    momentum=0.1),
                act_cfg=dict(
                    # torch.nn.SiLU
                    type='SiLU', 
                    inplace=False)),
            dict(
                # mmpretrain.models.backbones.resnet.BasicBlock
                type='BasicBlock',
                in_channels=16,
                out_channels=16,
                expansion=1,
                stride=1,
                dilation=1,
                downsample=None,
                style='pytorch',
                with_cp=False,
                conv_cfg=None,
                norm_cfg=dict(
                    type='BatchNorm2d', 
                    eps=1e-05, 
                    momentum=0.1),
                drop_path_rate=0.15,
                act_cfg=dict(
                    type='SiLU', 
                    inplace=False)),
            dict(
                # nas_special_modules.ParallelWithConfig
                type='ParallelWithConfig',
                module_cfg1=dict(
                    # nas_special_modules.SequentialWithConfig
                    type='SequentialWithConfig',
                    module_cfgs=[
                        dict(
                            # torch.nn.Conv2d
                            type='Conv2d',
                            in_channels=16,
                            out_channels=16,
                            kernel_size=3,
                            stride=1,
                            padding=1,
                            dilation=1,
                            groups=1,
                            bias=True,
                            padding_mode='zeros',
                            device=None,
                            dtype=None),
                        dict(
                            type='BatchNorm2d',
                            num_features=16,
                            eps=0.001,
                            momentum=0.1,
                            affine=True,
                            track_running_stats=True,
                            device=None,
                            dtype=None),
                        dict(
                            type='SiLU', 
                            inplace=False),
                        dict(
                            # nas_special_modules.MyReshape
                            type='MyReshape', 
                            shape=[
                                -1,
                                16,
                                16,
                                16
                            ]),
                        dict(
                            # timm.layers.bottleneck_attn.BottleneckAttn
                            type='BottleneckAttn',
                            dim=16,
                            dim_out=16,
                            feat_size=(
                                16,
                                16,
                            ),
                            stride=1,
                            num_heads=4,
                            dim_head=None,
                            qk_ratio=1.0,
                            qkv_bias=False,
                            scale_pos_embed=True),
                        dict(
                            type='MyReshape', 
                            shape=[
                                -1,
                                16,
                                16,
                                16
                            ]),
                        dict(
                            type='Conv2d',
                            in_channels=16,
                            out_channels=16,
                            kernel_size=3,
                            stride=1,
                            padding=1,
                            dilation=1,
                            groups=1,
                            bias=True,
                            padding_mode='zeros',
                            device=None,
                            dtype=None),
                        dict(
                            type='BatchNorm2d',
                            num_features=16,
                            eps=0.001,
                            momentum=0.1,
                            affine=True,
                            track_running_stats=True,
                            device=None,
                            dtype=None),
                        dict(
                            type='SiLU', 
                            inplace=False),
                    ]),
                module_cfg2=dict(
                    type='SequentialWithConfig',
                    module_cfgs=[
                        dict(
                            type='Conv2d',
                            in_channels=16,
                            out_channels=16,
                            kernel_size=1,
                            stride=1,
                            padding=0,
                            dilation=1,
                            groups=1,
                            bias=True,
                            padding_mode='zeros',
                            device=None,
                            dtype=None),
                        dict(
                            type='BatchNorm2d',
                            num_features=16,
                            eps=0.001,
                            momentum=0.1,
                            affine=True,
                            track_running_stats=True,
                            device=None,
                            dtype=None),
                        dict(
                            type='SiLU', 
                            inplace=False),
                    ]),
                merge_operation='add',
                concat_dim=1),
            dict(
                type='BasicBlock',
                in_channels=16,
                out_channels=16,
                expansion=1,
                stride=1,
                dilation=1,
                downsample=None,
                style='pytorch',
                with_cp=False,
                conv_cfg=None,
                norm_cfg=dict(
                    type='BatchNorm2d', 
                    eps=1e-05, 
                    momentum=0.1),
                drop_path_rate=0.15,
                act_cfg=dict(
                    type='SiLU', 
                    inplace=False)),
            dict(
                type='BottleneckAttn',
                dim=16,
                dim_out=None,
                feat_size=(
                    16,
                    16,
                ),
                stride=1,
                num_heads=4,
                dim_head=None,
                qk_ratio=1.0,
                qkv_bias=False,
                scale_pos_embed=True),
            dict(
                type='ParallelWithConfig',
                module_cfg1=dict(
                    type='SequentialWithConfig',
                    module_cfgs=[
                        dict(
                            type='Conv2d',
                            in_channels=16,
                            out_channels=32,
                            kernel_size=3,
                            stride=1,
                            padding=1,
                            dilation=1,
                            groups=1,
                            bias=True,
                            padding_mode='zeros',
                            device=None,
                            dtype=None),
                        dict(
                            type='BatchNorm2d',
                            num_features=32,
                            eps=1e-05,
                            momentum=0.1,
                            affine=True,
                            track_running_stats=True,
                            device=None,
                            dtype=None),
                        dict(
                            type='SiLU', 
                            inplace=False),
                        dict(
                            # timm.layers.mlp.ConvMlp
                            type='ConvMlp',
                            in_features=32,
                            hidden_features=64,
                            out_features=32,
                            norm_cfg=dict(
                                type='BatchNorm2d', 
                                eps=0.001, 
                                momentum=0.1),
                            act_cfg=dict(
                                # torch.nn.GELU
                                type='GELU'),
                            drop=0.1,
                            init_cfg=None),
                        dict(
                            type='BatchNorm2d',
                            num_features=32,
                            eps=1e-05,
                            momentum=0.1,
                            affine=True,
                            track_running_stats=True,
                            device=None,
                            dtype=None),
                        dict(
                            # timm.models.repvit.ConvNorm
                            type='ConvNorm_4',
                            in_dim=32,
                            out_dim=32,
                            ks=3,
                            stride=1,
                            pad=1,
                            dilation=1,
                            groups=1,
                            bn_weight_init=1),
                        dict(
                            type='ConvNorm_4',
                            in_dim=32,
                            out_dim=32,
                            ks=3,
                            stride=1,
                            pad=1,
                            dilation=1,
                            groups=1,
                            bn_weight_init=1),
                        dict(
                            type='SiLU', 
                            inplace=False),
                        dict(
                            type='Conv2d',
                            in_channels=32,
                            out_channels=16,
                            kernel_size=3,
                            stride=1,
                            padding=1,
                            dilation=1,
                            groups=1,
                            bias=True,
                            padding_mode='zeros',
                            device=None,
                            dtype=None),
                        dict(
                            type='BatchNorm2d',
                            num_features=16,
                            eps=1e-05,
                            momentum=0.1,
                            affine=True,
                            track_running_stats=True,
                            device=None,
                            dtype=None),
                        dict(
                            type='SiLU', 
                            inplace=False),
                    ]),
                module_cfg2=dict(
                    type='SequentialWithConfig',
                    module_cfgs=[
                        dict(
                            type='Conv2d',
                            in_channels=16,
                            out_channels=16,
                            kernel_size=1,
                            stride=1,
                            padding=0,
                            dilation=1,
                            groups=1,
                            bias=True,
                            padding_mode='zeros',
                            device=None,
                            dtype=None),
                        dict(
                            type='BatchNorm2d',
                            num_features=16,
                            eps=1e-05,
                            momentum=0.1,
                            affine=True,
                            track_running_stats=True,
                            device=None,
                            dtype=None),
                        dict(
                            type='SiLU', 
                            inplace=False),
                    ]),
                merge_operation='add',
                concat_dim=1),
            dict(
                type='BasicBlock',
                in_channels=16,
                out_channels=16,
                expansion=1,
                stride=1,
                dilation=1,
                downsample=None,
                style='pytorch',
                with_cp=False,
                conv_cfg=None,
                norm_cfg=dict(
                    type='BatchNorm2d', 
                    eps=1e-05, 
                    momentum=0.1),
                drop_path_rate=0.15,
                act_cfg=dict(
                    type='SiLU', 
                    inplace=False)),
            dict(
                # mmpretrain.models.utils.inverted_residual.InvertedResidual
                type='InvertedResidual_3',
                in_channels=16,
                out_channels=32,
                mid_channels=256,
                kernel_size=3,
                stride=2,
                drop_path_rate=0.2,
                se_cfg=None,
                norm_cfg=dict(
                    type='BatchNorm2d', 
                    eps=1e-05, 
                    momentum=0.1),
                act_cfg=dict(type='SiLU')),
            dict(
                type='BasicBlock',
                in_channels=32,
                out_channels=32,
                expansion=1,
                stride=1,
                dilation=1,
                downsample=None,
                style='pytorch',
                with_cp=False,
                conv_cfg=None,
                norm_cfg=dict(
                    type='BatchNorm2d', 
                    eps=1e-05, 
                    momentum=0.1),
                drop_path_rate=0.15,
                act_cfg=dict(
                    type='SiLU', 
                    inplace=False)),
            dict(
                type='BasicBlock',
                in_channels=32,
                out_channels=32,
                expansion=1,
                stride=1,
                dilation=1,
                downsample=None,
                style='pytorch',
                with_cp=False,
                conv_cfg=None,
                norm_cfg=dict(
                    type='BatchNorm2d', 
                    eps=1e-05, 
                    momentum=0.1),
                drop_path_rate=0.15,
                act_cfg=dict(
                    type='SiLU', 
                    inplace=False)),
            dict(
                type='BasicBlock',
                in_channels=32,
                out_channels=32,
                expansion=1,
                stride=1,
                dilation=1,
                downsample=None,
                style='pytorch',
                with_cp=False,
                conv_cfg=None,
                norm_cfg=dict(
                    type='BatchNorm2d', 
                    eps=1e-05, 
                    momentum=0.1),
                drop_path_rate=0.15,
                act_cfg=dict(
                    type='SiLU', 
                    inplace=False)),
            dict(
                type='BasicBlock',
                in_channels=32,
                out_channels=32,
                expansion=1,
                stride=1,
                dilation=1,
                downsample=None,
                style='pytorch',
                with_cp=False,
                conv_cfg=None,
                norm_cfg=dict(
                    type='BatchNorm2d', 
                    eps=1e-05, 
                    momentum=0.1),
                drop_path_rate=0.15,
                act_cfg=dict(
                    type='SiLU', 
                    inplace=False)),
            dict(
                type='BottleneckAttn',
                dim=32,
                dim_out=None,
                feat_size=(
                    8,
                    8,
                ),
                stride=1,
                num_heads=4,
                dim_head=None,
                qk_ratio=1.0,
                qkv_bias=False,
                scale_pos_embed=True),
            dict(
                type='BasicBlock',
                in_channels=32,
                out_channels=32,
                expansion=1,
                stride=1,
                dilation=1,
                downsample=None,
                style='pytorch',
                with_cp=False,
                conv_cfg=None,
                norm_cfg=dict(
                    type='BatchNorm2d', 
                    eps=1e-05, 
                    momentum=0.1),
                drop_path_rate=0.15,
                act_cfg=dict(
                    type='SiLU', 
                    inplace=False)),
            dict(
                type='BasicBlock',
                in_channels=32,
                out_channels=32,
                expansion=1,
                stride=1,
                dilation=1,
                downsample=None,
                style='pytorch',
                with_cp=False,
                conv_cfg=None,
                norm_cfg=dict(
                    type='BatchNorm2d', 
                    eps=1e-05, 
                    momentum=0.1),
                drop_path_rate=0.15,
                act_cfg=dict(
                    type='SiLU', 
                    inplace=False)),
            dict(
                type='BasicBlock',
                in_channels=32,
                out_channels=32,
                expansion=1,
                stride=1,
                dilation=1,
                downsample=None,
                style='pytorch',
                with_cp=False,
                conv_cfg=None,
                norm_cfg=dict(
                    type='BatchNorm2d', 
                    eps=1e-05, 
                    momentum=0.1),
                drop_path_rate=0.15,
                act_cfg=dict(
                    type='SiLU', 
                    inplace=False)),
            dict(
                type='BasicBlock',
                in_channels=32,
                out_channels=32,
                expansion=1,
                stride=1,
                dilation=1,
                downsample=None,
                style='pytorch',
                with_cp=False,
                conv_cfg=None,
                norm_cfg=dict(
                    type='BatchNorm2d', 
                    eps=1e-05, 
                    momentum=0.1),
                drop_path_rate=0.15,
                act_cfg=dict(
                    type='SiLU', 
                    inplace=False)),
            dict(
                type='InvertedResidual_3',
                in_channels=32,
                out_channels=64,
                mid_channels=320,
                kernel_size=3,
                stride=2,
                drop_path_rate=0.2,
                se_cfg=None,
                norm_cfg=dict(
                    type='BatchNorm2d', 
                    eps=1e-05, 
                    momentum=0.1),
                act_cfg=dict(type='SiLU')),
            dict(
                type='BasicBlock',
                in_channels=64,
                out_channels=64,
                expansion=1,
                stride=1,
                dilation=1,
                downsample=None,
                style='pytorch',
                with_cp=False,
                conv_cfg=None,
                norm_cfg=dict(
                    type='BatchNorm2d', 
                    eps=1e-05, 
                    momentum=0.1),
                drop_path_rate=0.15,
                act_cfg=dict(
                    type='SiLU', 
                    inplace=False)),
            dict(
                type='BasicBlock',
                in_channels=64,
                out_channels=64,
                expansion=1,
                stride=1,
                dilation=1,
                downsample=None,
                style='pytorch',
                with_cp=False,
                conv_cfg=None,
                norm_cfg=dict(
                    type='BatchNorm2d', 
                    eps=1e-05, 
                    momentum=0.1),
                drop_path_rate=0.15,
                act_cfg=dict(
                    type='SiLU', 
                    inplace=False)),
            dict(
                type='BasicBlock',
                in_channels=64,
                out_channels=64,
                expansion=1,
                stride=1,
                dilation=1,
                downsample=None,
                style='pytorch',
                with_cp=False,
                conv_cfg=None,
                norm_cfg=dict(
                    type='BatchNorm2d', 
                    eps=1e-05, 
                    momentum=0.1),
                drop_path_rate=0.15,
                act_cfg=dict(
                    type='SiLU', 
                    inplace=False)),
            dict(
                type='BasicBlock',
                in_channels=64,
                out_channels=64,
                expansion=1,
                stride=1,
                dilation=1,
                downsample=None,
                style='pytorch',
                with_cp=False,
                conv_cfg=None,
                norm_cfg=dict(
                    type='BatchNorm2d', 
                    eps=1e-05, 
                    momentum=0.1),
                drop_path_rate=0.15,
                act_cfg=dict(
                    type='SiLU', 
                    inplace=False)),
            dict(
                type='BasicBlock',
                in_channels=64,
                out_channels=64,
                expansion=1,
                stride=1,
                dilation=1,
                downsample=None,
                style='pytorch',
                with_cp=False,
                conv_cfg=None,
                norm_cfg=dict(
                    type='BatchNorm2d', 
                    eps=1e-05, 
                    momentum=0.1),
                drop_path_rate=0.15,
                act_cfg=dict(
                    type='SiLU', 
                    inplace=False)),
        ]),
    neck=dict(
        # mmpretrain.models.necks.gap.GlobalAveragePooling
        type='GlobalAveragePooling'),
    head=dict(
        # mmpretrain.models.heads.linear_head.LinearClsHead
        type='LinearClsHead',
        in_channels=64,
        loss=dict(
            # mmpretrain.models.losses.cross_entropy_loss.CrossEntropyLoss
            type='CrossEntropyLoss_2',
            loss_weight=1.0),
        num_classes=120,
        topk=(
            1,
            5,
        )))
\end{tcblisting}

\section{CIFAR-10}

\begin{tcblisting}{boxrule=2pt, listing only,listing options={language=iPython},title={Discovered architecture for CIFAR-10 \\(Acc. 95.31\%, 186 MFLOPs, 1.21 \#Params[M])}, breakable}
model = dict(
    # mmpretrain.models.classifiers.image.ImageClassifier
    type='ImageClassifier',
    backbone=dict(
        # nas_special_modules.NAS_Backbone
        type='NAS_Backbone', 
        layer_cfgs=[
            dict(
                # mmcv.cnn.bricks.conv_module.ConvModule
                type='ConvModule', 
                in_channels=3,
                out_channels=16,
                kernel_size=7,
                stride=1,
                padding=3,
                dilation=1,
                groups=1,
                bias=False,
                with_spectral_norm=False,
                order=(
                    'conv',
                    'norm',
                    'act',
                ),
                norm_cfg=dict(
                    # torch.nn.BatchNorm2d
                    type='BatchNorm2d', 
                    eps=0.001, 
                    momentum=0.1),
                act_cfg=dict(
                    # timm.layers.activations_me.SwishMe
                    type='SwishMe', 
                    inplace=False)),
            dict(
                # nas_special_modules.ParallelWithConfig
                type='ParallelWithConfig', 
                module_cfg1=dict(
                    # nas_special_modules.SequentialWithConfig
                    type='SequentialWithConfig', 
                    module_cfgs=[
                        dict(
                            # torch.nn.Conv2d
                            type='Conv2d', 
                            in_channels=16,
                            out_channels=32,
                            kernel_size=1,
                            stride=1,
                            padding=0,
                            dilation=1,
                            groups=1,
                            bias=True,
                            padding_mode='zeros',
                            device=None,
                            dtype=None),
                        dict(
                            # torch.nn.GELU
                            type='GELU', 
                            approximate='tanh'),
                        dict(
                            # timm.models.swiftformer.
                            # EfficientAdditiveAttention
                            type='EfficientAdditiveAttention', 
                            in_dims=32,
                            token_dim=16,
                            num_heads=1),
                    ]),
                module_cfg2=dict(
                    type='SequentialWithConfig',
                    module_cfgs=[
                        dict(
                            type='Conv2d',
                            in_channels=16,
                            out_channels=16,
                            kernel_size=3,
                            stride=1,
                            padding=1,
                            dilation=1,
                            groups=1,
                            bias=True,
                            padding_mode='zeros',
                            device=None,
                            dtype=None),
                        dict(
                            type='GELU', 
                            approximate='tanh'),
                        dict(
                            type='Conv2d',
                            in_channels=16,
                            out_channels=16,
                            kernel_size=1,
                            stride=1,
                            padding=0,
                            dilation=1,
                            groups=1,
                            bias=True,
                            padding_mode='zeros',
                            device=None,
                            dtype=None),
                    ]),
                merge_operation='mul',
                concat_dim=1),
            dict(
                # mmcv.cnn.bricks.context_block.ContextBlock
                type='ContextBlock', 
                in_channels=16,
                ratio=0.5,
                pooling_type='att',
                fusion_types=[
                    'channel_add',
                    'channel_mul',
                ]),
            dict(
                # mmpretrain.models.backbones.resnet.BasicBlock
                type='BasicBlock', 
                in_channels=16,
                out_channels=16,
                expansion=1,
                stride=1,
                dilation=1,
                downsample=None,
                style='pytorch',
                with_cp=False,
                conv_cfg=None,
                norm_cfg=dict(
                    type='BatchNorm2d',
                    eps=0.001, 
                    momentum=0.1),
                drop_path_rate=0.15,
                act_cfg=dict(
                    type='SwishMe', 
                    inplace=False)),
            dict(
                type='BasicBlock',
                in_channels=16,
                out_channels=16,
                expansion=1,
                stride=1,
                dilation=1,
                downsample=None,
                style='pytorch',
                with_cp=False,
                conv_cfg=None,
                norm_cfg=dict(
                    type='BatchNorm2d', 
                    eps=0.001, 
                    momentum=0.1),
                drop_path_rate=0.15,
                act_cfg=dict(
                    type='SwishMe', 
                    inplace=False)),
            dict(
                type='BasicBlock',
                in_channels=16,
                out_channels=16,
                expansion=1,
                stride=1,
                dilation=1,
                downsample=None,
                style='pytorch',
                with_cp=False,
                conv_cfg=None,
                norm_cfg=dict(
                    type='BatchNorm2d', 
                    eps=0.001, 
                    momentum=0.1),
                drop_path_rate=0.15,
                act_cfg=dict(
                    type='SwishMe', 
                    inplace=False)),
            dict(
                type='BasicBlock',
                in_channels=16,
                out_channels=16,
                expansion=1,
                stride=1,
                dilation=1,
                downsample=None,
                style='pytorch',
                with_cp=False,
                conv_cfg=None,
                norm_cfg=dict(
                    type='BatchNorm2d', 
                    eps=0.001, 
                    momentum=0.1),
                drop_path_rate=0.15,
                act_cfg=dict(
                    type='SwishMe', 
                    inplace=False)),
            dict(
                # mmpretrain.models.utils.inverted_residual.
                # InvertedResidual
                type='InvertedResidual_3', 
                in_channels=16,
                out_channels=32,
                mid_channels=1536,
                kernel_size=3,
                stride=2,
                drop_path_rate=0.3,
                se_cfg=None,
                norm_cfg=dict(
                    type='BatchNorm2d', 
                    eps=0.001, 
                    momentum=0.1),
                act_cfg=dict(
                    type='SwishMe', 
                    inplace=False)),
            dict(
                type='BasicBlock',
                in_channels=32,
                out_channels=32,
                expansion=1,
                stride=1,
                dilation=1,
                downsample=None,
                style='pytorch',
                with_cp=False,
                conv_cfg=None,
                norm_cfg=dict(
                    type='BatchNorm2d', 
                    eps=0.001, 
                    momentum=0.1),
                drop_path_rate=0.2,
                act_cfg=dict(
                    type='SwishMe', 
                    inplace=False)),
            dict(
                type='BasicBlock',
                in_channels=32,
                out_channels=32,
                expansion=1,
                stride=1,
                dilation=1,
                downsample=None,
                style='pytorch',
                with_cp=False,
                conv_cfg=None,
                norm_cfg=dict(
                    type='BatchNorm2d', 
                    eps=0.001, 
                    momentum=0.1),
                drop_path_rate=0.2,
                act_cfg=dict(
                    type='SwishMe', 
                    inplace=False)),
            dict(
                type='BasicBlock',
                in_channels=32,
                out_channels=32,
                expansion=1,
                stride=1,
                dilation=1,
                downsample=None,
                style='pytorch',
                with_cp=False,
                conv_cfg=None,
                norm_cfg=dict(
                    type='BatchNorm2d', 
                    eps=0.001, 
                    momentum=0.1),
                drop_path_rate=0.2,
                act_cfg=dict(
                    type='SwishMe', 
                    inplace=False)),
            dict(
                type='BasicBlock',
                in_channels=32,
                out_channels=32,
                expansion=1,
                stride=1,
                dilation=1,
                downsample=None,
                style='pytorch',
                with_cp=False,
                conv_cfg=None,
                norm_cfg=dict(
                    type='BatchNorm2d', 
                    eps=0.001, 
                    momentum=0.1),
                drop_path_rate=0.2,
                act_cfg=dict(
                    type='SwishMe', 
                    inplace=False)),
            dict(
                type='InvertedResidual_3',
                in_channels=32,
                out_channels=64,
                mid_channels=2048,
                kernel_size=3,
                stride=2,
                drop_path_rate=0.4,
                se_cfg=None,
                norm_cfg=dict(
                    type='BatchNorm2d', 
                    eps=0.001, 
                    momentum=0.1),
                act_cfg=dict(
                    type='SwishMe', 
                    inplace=False)),
            dict(
                type='BasicBlock',
                in_channels=64,
                out_channels=64,
                expansion=1,
                stride=1,
                dilation=1,
                downsample=None,
                style='pytorch',
                with_cp=False,
                conv_cfg=None,
                norm_cfg=dict(
                    type='BatchNorm2d', 
                    eps=0.001, 
                    momentum=0.1),
                drop_path_rate=0.25,
                act_cfg=dict(
                    type='SwishMe', 
                    inplace=False)),
            dict(
                type='BasicBlock',
                in_channels=64,
                out_channels=64,
                expansion=1,
                stride=1,
                dilation=1,
                downsample=None,
                style='pytorch',
                with_cp=False,
                conv_cfg=None,
                norm_cfg=dict(
                    type='BatchNorm2d', 
                    eps=0.001, 
                    momentum=0.1),
                drop_path_rate=0.25,
                act_cfg=dict(
                    type='SwishMe', 
                    inplace=False)),
            dict(
                # mmpretrain.models.backbones.vig.FFN
                type='FFN', 
                in_features=64,
                hidden_features=1024,
                out_features=64,
                act_cfg=dict(type='SwishMe', inplace=False),
                drop_path=0.25),
            dict(
                type='BasicBlock',
                in_channels=64,
                out_channels=64,
                expansion=1,
                stride=1,
                dilation=1,
                downsample=None,
                style='pytorch',
                with_cp=False,
                conv_cfg=None,
                norm_cfg=dict(
                    type='BatchNorm2d', 
                    eps=0.001, 
                    momentum=0.1),
                drop_path_rate=0.25,
                act_cfg=dict(
                    type='SwishMe', 
                    inplace=False)),
            dict(
                type='BasicBlock',
                in_channels=64,
                out_channels=64,
                expansion=1,
                stride=1,
                dilation=1,
                downsample=None,
                style='pytorch',
                with_cp=False,
                conv_cfg=None,
                norm_cfg=dict(
                    type='BatchNorm2d', 
                    eps=0.001, 
                    momentum=0.1),
                drop_path_rate=0.25,
                act_cfg=dict(
                    type='SwishMe', 
                    inplace=False)),
            dict(
                type='BasicBlock',
                in_channels=64,
                out_channels=64,
                expansion=1,
                stride=1,
                dilation=1,
                downsample=None,
                style='pytorch',
                with_cp=False,
                conv_cfg=None,
                norm_cfg=dict(
                    type='BatchNorm2d', 
                    eps=0.001, 
                    momentum=0.1),
                drop_path_rate=0.25,
                act_cfg=dict(
                    type='SwishMe', 
                    inplace=False)),
            dict(
                # mmpretrain.models.backbones.tinyvit.MBConvBlock
                type='MBConvBlock', 
                in_channels=64,
                out_channels=64,
                expand_ratio=2,
                drop_path=0.25,
                act_cfg=dict(
                    type='SwishMe', 
                    inplace=False)),
            dict(
                type='MBConvBlock',
                in_channels=64,
                out_channels=64,
                expand_ratio=2,
                drop_path=0.25,
                act_cfg=dict(
                    type='SwishMe', 
                    inplace=False)),
            dict(
                type='MBConvBlock',
                in_channels=64,
                out_channels=64,
                expand_ratio=4,
                drop_path=0.3,
                act_cfg=dict(
                    type='SwishMe', 
                    inplace=False)),
            dict(
                type='MBConvBlock',
                in_channels=64,
                out_channels=64,
                expand_ratio=4,
                drop_path=0.3,
                act_cfg=dict(
                    type='SwishMe', 
                    inplace=False)),
            dict(
                type='MBConvBlock',
                in_channels=64,
                out_channels=64,
                expand_ratio=4,
                drop_path=0.3,
                act_cfg=dict(
                    type='SwishMe', 
                    inplace=False)),
            dict(
                type='MBConvBlock',
                in_channels=64,
                out_channels=64,
                expand_ratio=6,
                drop_path=0.35,
                act_cfg=dict(
                    type='SwishMe', 
                    inplace=False)),
            dict(
                type='MBConvBlock',
                in_channels=64,
                out_channels=64,
                expand_ratio=6,
                drop_path=0.35,
                act_cfg=dict(
                    type='SwishMe', 
                    inplace=False)),
            dict(
                type='MBConvBlock',
                in_channels=64,
                out_channels=64,
                expand_ratio=6,
                drop_path=0.35,
                act_cfg=dict(
                    type='SwishMe', 
                    inplace=False)),
            dict(
                type='MBConvBlock',
                in_channels=64,
                out_channels=64,
                expand_ratio=8,
                drop_path=0.5,
                act_cfg=dict(
                    type='SwishMe', 
                    inplace=False)),
            dict(
                type='MBConvBlock',
                in_channels=64,
                out_channels=64,
                expand_ratio=8,
                drop_path=0.5,
                act_cfg=dict(
                    type='SwishMe', 
                    inplace=False)),
            dict(
                type='MBConvBlock',
                in_channels=64,
                out_channels=64,
                expand_ratio=8,
                drop_path=0.5,
                act_cfg=dict(
                    type='SwishMe', 
                    inplace=False)),
        ]),
    neck=dict(
        # mmpretrain.models.necks.gap.GlobalAveragePooling
        type='GlobalAveragePooling' 
    ),
    head=dict(
        # mmpretrain.models.heads.linear_head.LinearClsHead
        type='LinearClsHead', 
        in_channels=64,
        loss=dict( 
            # mmpretrain.models.losses.cross_entropy_loss.
            # CrossEntropyLoss
            type='CrossEntropyLoss_2', 
            loss_weight=1.0),
        num_classes=100,
        topk=(
            1,
            5,
        )))
\end{tcblisting}

\section{CIFAR-100}

\begin{tcblisting}{boxrule=2pt, listing only,listing options={language=iPython},title={Discovered architecture for CIFAR-100 \\(Acc. 77.83\%, 184 MFLOPs, 1.45 \#Params[M])}, breakable}
model = dict(
    # mmpretrain.models.classifiers.image.ImageClassifier
    type='ImageClassifier',
    backbone=dict(
        # nas_special_modules.NAS_Backbone
        type='NAS_Backbone', 
        layer_cfgs=[
            dict(
                # mmcv.cnn.bricks.conv_module.ConvModule
                type='ConvModule', 
                in_channels=3,
                out_channels=16,
                kernel_size=7,
                stride=1,
                padding=3,
                dilation=1,
                groups=1,
                bias=False,
                with_spectral_norm=False,
                order=(
                    'conv',
                    'norm',
                    'act',
                ),
                norm_cfg=dict(
                    # torch.nn.BatchNorm2d
                    type='BatchNorm2d', 
                    eps=0.001, 
                    momentum=0.1),
                act_cfg=dict(
                    # timm.layers.activations_me.SwishMe
                    type='SwishMe', 
                    inplace=False)),
            dict(
                # nas_special_modules.ParallelWithConfig
                type='ParallelWithConfig', 
                module_cfg1=dict(
                    # nas_special_modules.SequentialWithConfig
                    type='SequentialWithConfig', 
                    module_cfgs=[
                        dict(
                            # torch.nn.Conv2d
                            type='Conv2d', 
                            in_channels=16,
                            out_channels=32,
                            kernel_size=1,
                            stride=1,
                            padding=0,
                            dilation=1,
                            groups=1,
                            bias=True,
                            padding_mode='zeros',
                            device=None,
                            dtype=None),
                        dict(
                            # torch.nn.GELU
                            type='GELU', 
                            approximate='tanh'),
                        dict(
                            # timm.models.swiftformer.
                            # EfficientAdditiveAttention
                            type='EfficientAdditiveAttention', 
                            in_dims=32,
                            token_dim=16,
                            num_heads=1),
                    ]),
                module_cfg2=dict(
                    type='SequentialWithConfig',
                    module_cfgs=[
                        dict(
                            type='Conv2d',
                            in_channels=16,
                            out_channels=16,
                            kernel_size=3,
                            stride=1,
                            padding=1,
                            dilation=1,
                            groups=1,
                            bias=True,
                            padding_mode='zeros',
                            device=None,
                            dtype=None),
                        dict(
                            type='GELU', 
                            approximate='tanh'),
                        dict(
                            type='Conv2d',
                            in_channels=16,
                            out_channels=16,
                            kernel_size=1,
                            stride=1,
                            padding=0,
                            dilation=1,
                            groups=1,
                            bias=True,
                            padding_mode='zeros',
                            device=None,
                            dtype=None),
                    ]),
                merge_operation='mul',
                concat_dim=1),
            dict(
                # mmcv.cnn.bricks.context_block.ContextBlock
                type='ContextBlock', 
                in_channels=16,
                ratio=0.5,
                pooling_type='att',
                fusion_types=[
                    'channel_add',
                    'channel_mul',
                ]),
            dict(
                # mmpretrain.models.backbones.resnet.BasicBlock
                type='BasicBlock', 
                in_channels=16,
                out_channels=16,
                expansion=1,
                stride=1,
                dilation=1,
                downsample=None,
                style='pytorch',
                with_cp=False,
                conv_cfg=None,
                norm_cfg=dict(
                    type='BatchNorm2d',
                    eps=0.001, 
                    momentum=0.1),
                drop_path_rate=0.15,
                act_cfg=dict(
                    type='SwishMe', 
                    inplace=False)),
            dict(
                type='BasicBlock',
                in_channels=16,
                out_channels=16,
                expansion=1,
                stride=1,
                dilation=1,
                downsample=None,
                style='pytorch',
                with_cp=False,
                conv_cfg=None,
                norm_cfg=dict(
                    type='BatchNorm2d', 
                    eps=0.001, 
                    momentum=0.1),
                drop_path_rate=0.15,
                act_cfg=dict(
                    type='SwishMe', 
                    inplace=False)),
            dict(
                type='BasicBlock',
                in_channels=16,
                out_channels=16,
                expansion=1,
                stride=1,
                dilation=1,
                downsample=None,
                style='pytorch',
                with_cp=False,
                conv_cfg=None,
                norm_cfg=dict(
                    type='BatchNorm2d', 
                    eps=0.001, 
                    momentum=0.1),
                drop_path_rate=0.15,
                act_cfg=dict(
                    type='SwishMe', 
                    inplace=False)),
            dict(
                type='BasicBlock',
                in_channels=16,
                out_channels=16,
                expansion=1,
                stride=1,
                dilation=1,
                downsample=None,
                style='pytorch',
                with_cp=False,
                conv_cfg=None,
                norm_cfg=dict(
                    type='BatchNorm2d', 
                    eps=0.001, 
                    momentum=0.1),
                drop_path_rate=0.15,
                act_cfg=dict(
                    type='SwishMe', 
                    inplace=False)),
            dict(
                # mmpretrain.models.utils.inverted_residual.
                # InvertedResidual
                type='InvertedResidual_3', 
                in_channels=16,
                out_channels=32,
                mid_channels=1536,
                kernel_size=3,
                stride=2,
                drop_path_rate=0.3,
                se_cfg=None,
                norm_cfg=dict(
                    type='BatchNorm2d', 
                    eps=0.001, 
                    momentum=0.1),
                act_cfg=dict(
                    type='SwishMe', 
                    inplace=False)),
            dict(
                type='BasicBlock',
                in_channels=32,
                out_channels=32,
                expansion=1,
                stride=1,
                dilation=1,
                downsample=None,
                style='pytorch',
                with_cp=False,
                conv_cfg=None,
                norm_cfg=dict(
                    type='BatchNorm2d', 
                    eps=0.001, 
                    momentum=0.1),
                drop_path_rate=0.2,
                act_cfg=dict(
                    type='SwishMe', 
                    inplace=False)),
            dict(
                type='BasicBlock',
                in_channels=32,
                out_channels=32,
                expansion=1,
                stride=1,
                dilation=1,
                downsample=None,
                style='pytorch',
                with_cp=False,
                conv_cfg=None,
                norm_cfg=dict(
                    type='BatchNorm2d', 
                    eps=0.001, 
                    momentum=0.1),
                drop_path_rate=0.2,
                act_cfg=dict(
                    type='SwishMe', 
                    inplace=False)),
            dict(
                type='BasicBlock',
                in_channels=32,
                out_channels=32,
                expansion=1,
                stride=1,
                dilation=1,
                downsample=None,
                style='pytorch',
                with_cp=False,
                conv_cfg=None,
                norm_cfg=dict(
                    type='BatchNorm2d', 
                    eps=0.001, 
                    momentum=0.1),
                drop_path_rate=0.2,
                act_cfg=dict(
                    type='SwishMe', 
                    inplace=False)),
            dict(
                type='BasicBlock',
                in_channels=32,
                out_channels=32,
                expansion=1,
                stride=1,
                dilation=1,
                downsample=None,
                style='pytorch',
                with_cp=False,
                conv_cfg=None,
                norm_cfg=dict(
                    type='BatchNorm2d', 
                    eps=0.001, 
                    momentum=0.1),
                drop_path_rate=0.2,
                act_cfg=dict(
                    type='SwishMe', 
                    inplace=False)),
            dict(
                type='InvertedResidual_3',
                in_channels=32,
                out_channels=64,
                mid_channels=2048,
                kernel_size=3,
                stride=2,
                drop_path_rate=0.4,
                se_cfg=None,
                norm_cfg=dict(
                    type='BatchNorm2d', 
                    eps=0.001, 
                    momentum=0.1),
                act_cfg=dict(
                    type='SwishMe', 
                    inplace=False)),
            dict(
                type='BasicBlock',
                in_channels=64,
                out_channels=64,
                expansion=1,
                stride=1,
                dilation=1,
                downsample=None,
                style='pytorch',
                with_cp=False,
                conv_cfg=None,
                norm_cfg=dict(
                    type='BatchNorm2d', 
                    eps=0.001, 
                    momentum=0.1),
                drop_path_rate=0.25,
                act_cfg=dict(
                    type='SwishMe', 
                    inplace=False)),
            dict(
                type='BasicBlock',
                in_channels=64,
                out_channels=64,
                expansion=1,
                stride=1,
                dilation=1,
                downsample=None,
                style='pytorch',
                with_cp=False,
                conv_cfg=None,
                norm_cfg=dict(
                    type='BatchNorm2d', 
                    eps=0.001, 
                    momentum=0.1),
                drop_path_rate=0.25,
                act_cfg=dict(
                    type='SwishMe', 
                    inplace=False)),
            dict(
                # mmpretrain.models.backbones.vig.FFN
                type='FFN', 
                in_features=64,
                hidden_features=1024,
                out_features=64,
                act_cfg=dict(type='SwishMe', inplace=False),
                drop_path=0.25),
            dict(
                type='BasicBlock',
                in_channels=64,
                out_channels=64,
                expansion=1,
                stride=1,
                dilation=1,
                downsample=None,
                style='pytorch',
                with_cp=False,
                conv_cfg=None,
                norm_cfg=dict(
                    type='BatchNorm2d', 
                    eps=0.001, 
                    momentum=0.1),
                drop_path_rate=0.25,
                act_cfg=dict(
                    type='SwishMe', 
                    inplace=False)),
            dict(
                type='BasicBlock',
                in_channels=64,
                out_channels=64,
                expansion=1,
                stride=1,
                dilation=1,
                downsample=None,
                style='pytorch',
                with_cp=False,
                conv_cfg=None,
                norm_cfg=dict(
                    type='BatchNorm2d', 
                    eps=0.001, 
                    momentum=0.1),
                drop_path_rate=0.25,
                act_cfg=dict(
                    type='SwishMe', 
                    inplace=False)),
            dict(
                type='BasicBlock',
                in_channels=64,
                out_channels=64,
                expansion=1,
                stride=1,
                dilation=1,
                downsample=None,
                style='pytorch',
                with_cp=False,
                conv_cfg=None,
                norm_cfg=dict(
                    type='BatchNorm2d', 
                    eps=0.001, 
                    momentum=0.1),
                drop_path_rate=0.25,
                act_cfg=dict(
                    type='SwishMe', 
                    inplace=False)),
            dict(
                # mmpretrain.models.backbones.tinyvit.MBConvBlock
                type='MBConvBlock', 
                in_channels=64,
                out_channels=64,
                expand_ratio=2,
                drop_path=0.25,
                act_cfg=dict(
                    type='SwishMe', 
                    inplace=False)),
            dict(
                type='MBConvBlock',
                in_channels=64,
                out_channels=64,
                expand_ratio=2,
                drop_path=0.25,
                act_cfg=dict(
                    type='SwishMe', 
                    inplace=False)),
            dict(
                type='MBConvBlock',
                in_channels=64,
                out_channels=64,
                expand_ratio=4,
                drop_path=0.3,
                act_cfg=dict(
                    type='SwishMe', 
                    inplace=False)),
            dict(
                type='MBConvBlock',
                in_channels=64,
                out_channels=64,
                expand_ratio=4,
                drop_path=0.3,
                act_cfg=dict(
                    type='SwishMe', 
                    inplace=False)),
            dict(
                type='MBConvBlock',
                in_channels=64,
                out_channels=64,
                expand_ratio=4,
                drop_path=0.3,
                act_cfg=dict(
                    type='SwishMe', 
                    inplace=False)),
            dict(
                type='MBConvBlock',
                in_channels=64,
                out_channels=64,
                expand_ratio=6,
                drop_path=0.35,
                act_cfg=dict(
                    type='SwishMe', 
                    inplace=False)),
            dict(
                type='MBConvBlock',
                in_channels=64,
                out_channels=64,
                expand_ratio=6,
                drop_path=0.35,
                act_cfg=dict(
                    type='SwishMe', 
                    inplace=False)),
            dict(
                type='MBConvBlock',
                in_channels=64,
                out_channels=64,
                expand_ratio=6,
                drop_path=0.35,
                act_cfg=dict(
                    type='SwishMe', 
                    inplace=False)),
            dict(
                type='MBConvBlock',
                in_channels=64,
                out_channels=64,
                expand_ratio=8,
                drop_path=0.5,
                act_cfg=dict(
                    type='SwishMe', 
                    inplace=False)),
            dict(
                type='MBConvBlock',
                in_channels=64,
                out_channels=64,
                expand_ratio=8,
                drop_path=0.5,
                act_cfg=dict(
                    type='SwishMe', 
                    inplace=False)),
            dict(
                type='MBConvBlock',
                in_channels=64,
                out_channels=64,
                expand_ratio=8,
                drop_path=0.5,
                act_cfg=dict(
                    type='SwishMe', 
                    inplace=False)),
        ]),
    neck=dict(
        # mmpretrain.models.necks.gap.GlobalAveragePooling
        type='GlobalAveragePooling' 
    ),
    head=dict(
        # mmpretrain.models.heads.linear_head.LinearClsHead
        type='LinearClsHead', 
        in_channels=64,
        loss=dict( 
            # mmpretrain.models.losses.cross_entropy_loss.
            # CrossEntropyLoss
            type='CrossEntropyLoss_2', 
            loss_weight=1.0),
        num_classes=100,
        topk=(
            1,
            5,
        )))
\end{tcblisting}

\end{document}